\documentclass{article}

\usepackage[nonatbib,final]{neurips_2022}
\usepackage[numbers]{natbib}

\usepackage[utf8]{inputenc} 
\usepackage[T1]{fontenc}    
\usepackage{hyperref}       
\usepackage{url}            
\usepackage{booktabs}       
\usepackage{amsfonts}       
\usepackage{amsmath}
\usepackage{nicefrac}       
\usepackage{microtype}      
\usepackage[table]{xcolor}  
\usepackage{graphicx}
\usepackage{multirow}
\usepackage{changepage,threeparttable} 
\usepackage{makecell}
\usepackage{colortbl}
\usepackage{xspace}
\usepackage{enumitem}
\usepackage{subfigure}
\usepackage{wrapfig}

\hypersetup{
    colorlinks,
    linkcolor={red!50!black},
    citecolor={blue!50!black},
    urlcolor={blue!80!black}
}

\title{Long Range Graph Benchmark}

\author{%
   Vijay Prakash Dwivedi\thanks{To whom correspondence should be addressed: \texttt{vijaypra001@e.ntu.edu.sg}} \\
   Nanyang Technological University, Singapore \\
   \And
   Ladislav Rampášek \\
   Mila, Université de Montréal \\
   \And
   Mikhail Galkin \\
   Mila, McGill University \\
   \And
   Ali Parviz \\
   NJIT, Mila, Université de Montréal \\
   \And
   Guy Wolf \\
   Mila, Université de Montréal \\
   \And
   Anh Tuan Luu \\
   \fontsize{9.7pt}{9.7pt}\selectfont{Nanyang Technological University, Singapore} \\
    \And
   Dominique Beaini \\
   \fontsize{9.7pt}{9.7pt}\selectfont{Valence Discovery, Mila, Université de Montréal} \\
}

\begin{document}


\newcommand{\pascal}{\texttt{PascalVOC-SP}\xspace}  
\newcommand{\coco}{\texttt{COCO-SP}\xspace}  
\newcommand{\pcqmcontact}{\texttt{PCQM-Contact}\xspace}
\newcommand{\pepfunc}{\texttt{Peptides-func}\xspace}
\newcommand{\pepstruct}{\texttt{Peptides-struct}\xspace}
\newcommand{\rbgraph}{\texttt{rag-boundary}\xspace}

\definecolor{lightgray}{gray}{0.95}
\definecolor{Gray}{gray}{0.85}
\definecolor{LightCyan}{rgb}{0.88,1,1}
\definecolor{LightPink}{HTML}{FCE1EF}
\definecolor{LightGreen}{HTML}{EEF7E1}
\newcolumntype{A}{>{\columncolor{white}}c}
\newcolumntype{B}{>{\columncolor{LightGreen}}c}
\newcolumntype{C}{>{\columncolor{LightPink}}c}

\definecolor{purpleheart}{rgb}{0.41, 0.21, 0.61}
\newcommand{\mycomment}[3]{\textcolor{#1}{[\textbf{#2:} #3]}}
\newcommand{\lr}[1]{\mycomment{orange}{LR}{#1}}
\newcommand{\db}[1]{\mycomment{magenta}{DB}{#1}}
\newcommand{\mg}[1]{\mycomment{green!70!black}{MG}{#1}}
\newcommand{\vd}[1]{\textcolor{purpleheart}{#1}} 
\newcommand{\draft}[1]{\textcolor{blue}{#1}}

\newcommand{\myparagraph}[1]{\noindent\textbf{#1}}

\definecolor{dark2green}{rgb}{0.1, 0.65, 0.3}
\definecolor{dark2orange}{rgb}{0.9, 0.4, 0.}
\definecolor{dark2purple}{rgb}{0.4, 0.4, 0.8}
\newcommand{\first}[1]{\textbf{#1}}
\newcommand{\second}[1]{#1}

\maketitle
\setcounter{footnote}{0}

\begin{abstract}
Graph Neural Networks (GNNs) that are based on the message passing (MP) paradigm generally exchange information between 1-hop neighbors to build node representations at each layer. In principle, such networks are not able to capture long-range interactions (LRI) that may be desired or necessary for learning a given task on graphs. Recently, there has been an increasing interest in development of Transformer-based methods for graphs that can consider full node connectivity beyond the original sparse structure, thus enabling the modeling of LRI. However, MP-GNNs that simply rely on 1-hop message passing often fare better in several existing graph benchmarks when combined with positional feature representations, among other innovations, hence limiting the perceived utility and ranking of Transformer-like architectures. Here, we present the Long Range Graph Benchmark (LRGB)\footnote{Open-sourced at \url{https://github.com/vijaydwivedi75/lrgb} and deposited at \href{https://doi.org/10.5281/zenodo.6975830}{Zenodo}~\cite{dwivedi_vijay_prakash_2022_6975830}.} with 5 graph learning datasets: \pascal, \coco, \pcqmcontact, \pepfunc and \pepstruct that arguably require LRI reasoning to achieve strong performance in a given task. We benchmark both baseline GNNs and Graph Transformer networks to verify that the models which capture long-range dependencies perform significantly better on these tasks. Therefore, these datasets are suitable for benchmarking and exploration of MP-GNNs and Graph Transformer architectures that are intended to capture LRI. 
\end{abstract}

\section{Introduction}

Considering a graph 
as a collection of nodes 
where arbitrary relations between nodes are represented as edges,
there are numerous real-world instances with data structures where complex and irregular interactions among objects can be represented as edges where the objects themselves are denoted as nodes. This has led to a rapid rise of interest in the development of graph neural networks (GNNs) \cite{hamilton, ma2021deep, bronstein2021geometric, wu2022graph} for deep learning on geometric and graph domains.

The popularly used class of GNNs is based on the message passing paradigm \cite{gilmer2017neural_mpnn} where a node's feature representation, at each layer, is updated using a trainable function that receives and combines feature information from all its neighboring nodes. A network that has $L$ GNN layers, stacked sequentially, can iteratively aggregate feature information from up to $L$ hops for the update of a node's representation. If a long-range information to a node from its $L$-hop neighbor is needed for a task (say, for a large $L$), the same number of GNN layers is ideally required. However, with the increasing $L$ the size of the $L$-hop neighborhood grows exponentially and so does the amount of information that needs to be encoded into one vector by the network. This leads to `information oversquashing' as the message aggregation step continues to be iteratively applied at each layer in order to propagate the information \cite{alon2020bottleneck}. Consequently, such GNNs fail at capturing long-range dependencies as a significant amount of distant information may get lost due to the squashing.

In order to factor in distant information when message passing GNNs are used, Alon et al. \cite{alon2020bottleneck} used a fully connected graph at the final layer as an intuitive remedy. The primary rationale behind this approach is to enable each node in a graph to connect to every other node at some stage in the network to pass the information that otherwise would get squashed, thus breaking the bottleneck. Consequently, several recent works propose Graph Transformers that leverage full-connections among all nodes in the graph to capture long-range dependencies \cite{kreuzer2021rethinking, ying2021transformers, mialon2021graphit}. 

\begin{wrapfigure}{r}{0.27\textwidth}
    \vspace{-6pt}
    \centering
    \includegraphics[width=1\linewidth]{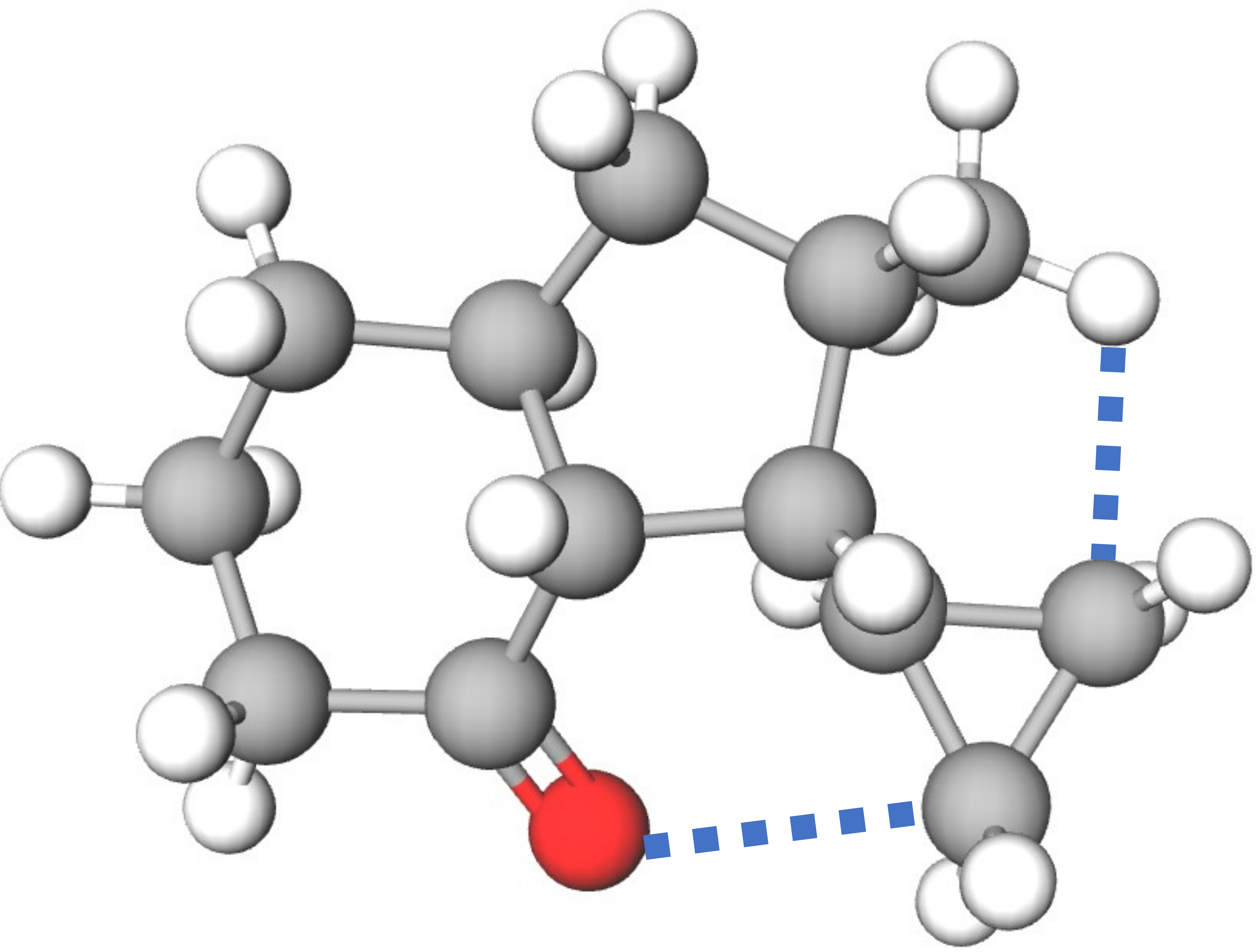}
    \label{fig:molecule-contact}
    \vspace{-15pt}
    \caption{
    Molecule with LRIs (dotted lines showing 3D atomic contact) that are not trivially captured by the graph structure.
    }
\end{wrapfigure}

However, it is often the case that these models are evaluated on datasets where the corresponding tasks primarily rely on local structural information rather than the distant information propagation between nodes. This observation is prevalent for many existing datasets such as ZINC \cite{dwivedi2020benchmarking}, \texttt{ogbg-molpcba}, or \texttt{ogbg-molhiv} \cite{hu2020open} that are among the most frequently used benchmarks. The Spectral Attention Network (SAN) \cite{kreuzer2021rethinking} has shown insignificant contribution of full attention in these benchmarks. In fact, leaderboards of these benchmarks are topped by local MP-GNN based models \cite{bodnar2021weisfeiler, zhang2021nested, lim2022sign}, albeit these GNNs are non trivial extensions and are augmented with higher-order structural information, among other model improvements. At the same time, these molecular benchmarks  largely consist of graphs of small sizes, \textit{i.e.}, the number of nodes in a graph. Nonetheless, on the contrary, graphs with large number of nodes may not necessarily imply that they require models with long-range dependencies for the learning task.

\begin{table}[!t]
    \caption{Overview of the datasets in the proposed LRGB. Note: `Pixels+Coord' denotes the feature vector consisting of 12-dim statistics of each superpixel (for each RGB color: the average, standard deviation, maximum, and minimum of pixel intensities in the superpixel) and 2-dim coordinates of the center of mass of its X,Y pixel locations. `Edge Weight' corresponds to the weight assigned between two superpixel nodes w.r.t. the construction method. The `Atom Encoder' and `Bond Encoder' are OGB molecular feature encoders \cite{hu2020open, hu2021ogb_lsc}. All tasks are inductive tasks.
    }\label{tab:data_overview}
    \begin{adjustwidth}{-2.5 cm}{-2.5 cm}\centering
    \setlength\tabcolsep{3.5pt} 
    \scalebox{0.8}{
    \begin{tabular}{l c c cc c}
    \toprule
    \textbf{Dataset} & \textbf{Domain} & \textbf{Task} & \textbf{Node Feat. (dim)} & \textbf{Edge Feat. (dim)} & \textbf{Perf. Metric}\\
      \midrule
    \pascal & \multirow{2}{*}{Computer Vision} & \multirow{2}{*}{Node Classif.} & \multirow{2}{*}{Pixel + Coord (14)} & \multirow{2}{*}{Edge Weight (1 or 2)} & \multirow{2}{*}{macro F1}\\
    \multirow{1}{*}{\coco} & & & & &\\
    \midrule
    \pcqmcontact & Quantum Chemistry & Link Prediction & Atom Encoder (9) & Bond Encoder (3) & Hits@K, MRR\\ 
    \midrule
    \pepfunc & \multirow{2}{*}{Chemistry} & Graph Classif. & \multirow{2}{*}{Atom Encoder (9)} & \multirow{2}{*}{Bond Encoder (3)} & AP\\
    \pepstruct & & Graph Regression & & & MAE\\
    \bottomrule
    \end{tabular}
    }
    \vspace{-5pt}
    \end{adjustwidth}
\end{table}

\begin{table}[t]
    \caption{Statistics of the five proposed LRGB datasets.
    }\label{tab:data_summary}
    \centering
    \setlength\tabcolsep{5pt} 
    \scalebox{0.82}{
    \begin{tabular}{l r rrr rr cc}
    \toprule
    \multirow{2}{*}{\textbf{Dataset}} & \multirow{2}{*}{\shortstack[c]{\textbf{Total} \\ \textbf{Graphs}}} & \multirow{2}{*}{\shortstack[c]{\textbf{Total} \\ \textbf{Nodes}}} & \multirow{2}{*}{\shortstack[c]{\textbf{Avg} \\ \textbf{Nodes}}} & 
    \multirow{2}{*}{\shortstack[c]{\textbf{Mean} \\\textbf{Deg.}}} &
    \multirow{2}{*}{\shortstack[c]{\textbf{Total} \\ \textbf{Edges}}} &
    \multirow{2}{*}{\shortstack[c]{\textbf{Avg} \\ \textbf{Edges}}} & 
    \multirow{2}{*}{\shortstack[c]{\textbf{Avg} \\ \textbf{Short.Path.}}} &
    \multirow{2}{*}{\shortstack[c]{\textbf{Avg} \\ \textbf{Diameter}}}\\
    & & & & & & & & \\\midrule
    \pascal & 11,355 & 5,443,545 & 479.40 & 5.65 & 30,777,444 & 2,710.48 & 10.74$\pm$0.51 & 27.62$\pm$2.13\\
    \coco & 123,286 & 58,793,216 & 476.88 & 5.65 & 332,091,902 & 2,693.67 & 10.66$\pm$0.55 & 27.39$\pm$2.14\\
    \pcqmcontact & 529,434 & 15,955,687 & 30.14 & 2.03 & 32,341,644 & 61.09 & 4.63$\pm$0.63 & 9.86$\pm$1.79\\
    \pepfunc& 15,535 & 2,344,859 & 150.94 & 2.04 & 4,773,974 & 307.30 & 20.89$\pm$9.79 & 56.99$\pm$28.72\\
    \pepstruct & 15,535 & 2,344,859 & 150.94 & 2.04 & 4,773,974 & 307.30 & 20.89$\pm$9.79 & 56.99$\pm$28.72\\
    \bottomrule
    \end{tabular}
    }
\end{table}

\textbf{Contribution.} In this work, we focus on these shortcomings of existing popular graph learning benchmark datasets and propose characterizing factors in a dataset that can be studied for the exploration of new GNN and Graph Transformer architectures that possess long-range interaction (LRI) capabilities. Note that our characterization henceforth is not directed at proposing `provable LRI' benchmarks, which would often lead to toy datasets (that are useful for quick prototyping of ideas)
such as the shortest path prediction task \cite{stachenfeld2020graph} or the color connectivity dataset \cite{rampavsek2021hierarchical} which rely on LRI. Instead, our aim is to propose real-world datasets that require LRI, and the factors we consider for a LRGB dataset characterization could be understood as implications which suggest that the learning task(s) in the graphs would depend on long range signal propagation.
Consequently, we introduce 5 benchmarking datasets -- \pascal, \coco, \pcqmcontact, \pepfunc and \pepstruct from the domains of Computer Vision and Chemistry which we incorporate in LRGB, see Tables \ref{tab:data_overview}-\ref{tab:data_summary} for an overview and Figure \ref{fig:molecule-contact} for a sample illustration. The learning tasks that we propose in these datasets depend on some degree of long-range signal handling given the nature of task, contribution of global graph structure to the task, and the sizes of graphs in these datasets. Fittingly, in our baseline experiments, these datasets show that the fully-connected models which enable LRI propagation perform considerably better than local message passing based GNNs.

\textbf{Existing attempts towards LRI benchmarks.} Thanks to the understanding of the limitations of message passing based GNNs \cite{gilmer2017neural_mpnn} with respect to the 1-Weisfeiler Leman (WL) isomorphism test \cite{weisfeiler1968reduction, xu2018powerful_gin, morris2019weisfeiler} and the information oversquashing \cite{alon2020bottleneck}, there has been several developments of GNN and Graph Transformer architectures which have strictly greater representation power than 1-WL \cite{bouritsas2022improving, bodnar2021weisfeiler, morris2019weisfeiler, morris2020weisfeiler, li_distance_2020, kreuzer2021rethinking, ying2021transformers}. 
By design, fully connected Graph Transformers \cite{ying2021transformers, kreuzer2021rethinking, mialon2021graphit} are able to model long-range dependencies in the graphs and alleviate information bottleneck to some extent \cite{shi2022benchmarking}. Similarly, some recent models have been designed to perform non-local feature integration in particular aspect of non-homophilic graphs \cite{pei2020geom, liu2021eignn}.
However, most of such architectures are evaluated on benchmarks where it is not clear whether long-range interactions are required for the corresponding learning tasks.
For natural language processing (NLP), the Long Range Arena \cite{tay2020long} benchmark has been instrumental in studying the capacity and efficiency of architectures against longer sequences.
Notably, a recent work has introduced non-homophilic graph datasets \cite{lim2021new} that is an orthogonal attempt to contribute towards graph learning testbeds beyond the widely used benchmarks which favor non-local GNN methods.

Nevertheless, we believe there is a consensus in the community towards the development of specific benchmarks that can assist 
LRI enabled-GNNs, including full-graph operable Graph Transformers. This can be observed in existing independent attempts at proposing new graph benchmarks to evaluate LRI. In Stachenfeld et al. \cite{stachenfeld2020graph}, a new Graph MNIST benchmark with an increased average graph diameter compared to MNIST superpixels \cite{defferrard2016convolutional, dwivedi2020benchmarking} was used to evaluate the ability of Spectral Graph Network in incorporating long range signals. Similarly, a synthetic color connectivity task that, by construction, requires LRI to differentiate between its classes was used in Rampášek \& Wolf \cite{rampavsek2021hierarchical} to demonstrate a hierarchical graph network's ability in modeling such signals. Another synthetic benchmark was used in Alon et al. \cite{alon2020bottleneck} to probe the oversquashing phenomenon and implement intuitive tricks such as a fully connected graph layer to alleviate the bottleneck. 
Finally, a Chains dataset \cite{gu2020implicit} was created for testing long-range dependency that was adopted to develop enhanced models such as higher order Transformers \cite{kim2021transformers}, among others \cite{yang2021graph, liu2022mgnni}.
Apart from the aforementioned synthetic and semi-real tasks, MalNet \cite{freitas2020large}, a real-world dataset of large function call graphs (avg. 15k nodes) was recently proposed. MalNet could be a potential LRI task given its graph sizes, which we discuss in Section~\ref{sec:characterizing} as a characterizing aspect of LRI benchmarks, along with other factors.

\section{Characterizing Long-Range Interactions}
\vspace{-3pt}
\label{sec:characterizing}
We now proceed to study the key characteristics that can help to determine how a graph dataset could be appropriate to guage whether a GNN can or cannot model LRIs. Note that our discussion here is based on datasets with inductive tasks, which contain many graphs, rather than a single (large) graph.

\textbf{Graph Size.} The number of nodes in a graph is critical to determine if any visible effect of information oversquashing would occur if a local message passing based GNN (MP-GNN) is used to learn on this graph. If $r$ is a hypothetical estimate of the learning problem's radius in the graph or the problem's range of interaction \cite{alon2020bottleneck}, and $L\geq r$ denotes the number of layers that are stacked in a GNN to learn the task, the number of nodes in a node's receptive field grows exponentially, \textit{i.e.}, $O(\text{exp}(L))$ \cite{alon2020bottleneck, chen2017stochastic}. However, if the graph size is small, such as \texttt{ogbg-mol*} \cite{hu2020open} or ZINC \cite{dwivedi2020benchmarking} datasets with average graph size in the range of 23-26 nodes, then $r$ would effectively be small as well. As a consequence, the effect of squashing of the information from the node's receptive field will be diminishing, and any local MP-GNN would succeed to learn the task to a great extent without being influenced by the information bottleneck. Therefore, a direct conclusion of this condition is that for a LRI benchmark, the graph sizes should be sufficiently large in order to separate the local MP-GNNs' performance from those models which model LRIs. However, this condition of graph size alone may not be enough to determine a LRGB dataset as the problem radius $r$ may be small for some tasks even if the graph size is large, which brings us to the following factors.

\textbf{Nature of Task.}
The nature of task can be understood to be directly related to the problem's range of interaction, $r$. In broad sense, the task can be \textit{either} short-range, \textit{i.e.}, requiring information exchange among nodes in local or near-local neighborhood, \textit{or} long-range, where interactions are required far away from the near-local neighborhood. For  instance, the task in the ZINC molecular dataset \cite{irwin2012zinc, dwivedi2020benchmarking} is associated with counting local structures and it has been revealed that a substructure-counting based model \cite{bouritsas2022improving} would optimally require counts of 7-length substructures for the best performance. Any increment above this length does not further show a gain in the ZINC task. It may therefore be interpreted that such a benchmarking task does not require long-range signal propagation. Additionally, note that the graph sizes in ZINC are small (9-37 nodes) which functionally makes it a non-LRI benchmark if we also factor in the nature of ZINC's task. 

However, even if the graph size of a dataset is considerably large, it may not warrant that models with long-range signal propagation are best suited unless the nature of the task determines so. A recent example to this is MalNet-Tiny dataset \cite{freitas2020large} consisting of graphs up to 5,000 nodes, where there is a scarce improvement of performance of fully-connected GNN modules \cite{rampasek2022GPS}.
Finally, there exist tasks in graphs which are prone to bottleneck if local MP-GNNs are used while this bottleneck is substantially reduced if LRI enabled non-local MP-GNNs are used, see Table 3 in Shi et al. \cite{shi2022benchmarking}.

\textbf{Contribution of global graph structure to task.}
Since MP-GNNs rely on information aggregated from a local neighborhood to update a node’s features, it is subject to miss global structural information, such as global positional encoding (PE) \cite{rampasek2022GPS}. Additionally, MP-GNNs are also susceptible to lose out critical node signals coming from distant nodes if the graph size is large enough \cite{alon2020bottleneck}. Such signals are conveniently propagated in a fully-connected Transformer-like networks modeling LRI. The contribution of global structure to a task thus becomes a distinctive property desired in a LRI benchmark. MP-GNNs are often augmented with positional encodings (PE) carrying global structural information to assist tasks requiring some degree of LRI. A dataset where the learning task benefits from global PE can hence be a potential LRI benchmark. Similarly, if the learning task in a dataset is dependent on some form of distance information, or is directly a function of distance, coupled with graph feature information, the dataset can be a strong candidate for LRGB since the distance information would require global structural information. Examples of this can be molecular datasets where the learning task is related to prediction of 2D or 3D distance and structure properties.

\section{Proposed LRGB Datasets}
\label{sec:proposed_datasets}

\subsection{\pascal}
\label{sec:pascalvoc}
\pascal is a node classification dataset, based on the Pascal VOC 2011 image dataset \cite{everingham2010pascal}, where each node corresponds to a region of the image belonging to a particular class. The original dataset is available on a Custom License (respecting Flickr terms of use) \cite{pascalvoc}. Similar to the recent superpixels (\texttt{SP}) datasets such as MNIST and CIFAR10 \cite{dwivedi2020benchmarking}, we extract superpixels nodes in \pascal by using the SLIC algorithm \cite{achanta2010slic} and construct a \rbgraph graph that interconnects these nodes.  
Unlike MNIST and CIFAR10 superpixels which have up to 75 and 150 nodes respectively, we extract a maximum of 500 superpixel nodes for SLIC compactness value of 30\footnote{The compactness parameter balances spatial and color information when extracting superpixels in SLIC \cite{achanta2010slic}.}
in \pascal in order to satisfy the `graph size’ characteristic to make it a LRGB benchmark. Effectively, it results in the \pascal dataset to have an average shortest path length of 10.74$\pm$0.51 and average diameter of 27.62$\pm$2.13 (see Table \ref{tab:data_summary}) which is significantly larger than that of MNIST with 3.03$\pm$0.17, 6.03$\pm$0.47 and CIFAR10 with 3.97$\pm$0.08, 8.46$\pm$0.50 average shortest path and diameters respectively. We argue that these properties, along with the task of predicting the node label of the superpixel region, which is analogous to the semantic segmentation task in Computer Vision, makes \pascal a suitable LRGB dataset fulfilling major characteristics discussed in Section \ref{sec:characterizing}. We also prepare other variants of \pascal with different values of SLIC compactness and graph construction options, which are included in Appendix \ref{app:dataset_details}.

\textbf{Statistics.} 
There are 11,355 graphs with a total of 5.4 million nodes in \pascal where each graph corresponds to an image in Pascal VOC 2011. The graphs prepared after the superpixels extraction have on average 479.40 nodes with complete statistics reported in Table \ref{tab:data_summary}.

\textbf{Task.} 
The 
task in \pascal is node classification which predicts a semantic segmentation label for each superpixel node out of 21 classes. We label each superpixel node with the same class label of the original pixel ground truth which is on the mean coordinates of the superpixel region.\footnote{The labels are based on the annotations provided in Semantic Boundary Dataset (SBD) version of Pascal VOC 2011: \fontsize{8.5pt}{8.5pt}\selectfont{\url{https://github.com/shelhamer/fcn.berkeleyvision.org/tree/master/data/pascal}}}

\textbf{Splitting.} In the original Pascal VOC 2011 dataset, there are only training and validation splits that we can use. For \pascal, we maintain the train set as it is, and split the original validation set into new validation and test sets. For this splitting, we divide the original validation set in 50:50 ratio using a stratified split proportionate to the original distribution of the data with respect to a meta label that depends on the node classes. This meta label is a ground truth class value obtained by a majority voting of non-background ground truth node labels. The splitting decision with this meta-label is taken to preserve a similar distribution of node labels in both the new validation and the test set. Thus, we have 8,498, 1,428 and 1,429 graphs in the final training, validation and test sets, respectively.

\begin{figure}[t]
    \centering
    \includegraphics[trim=150 20 125 10, clip, width=1\textwidth]{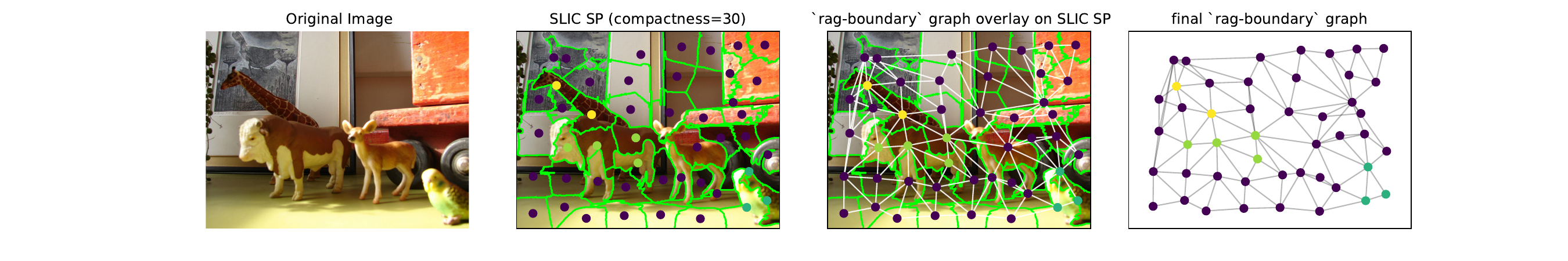}
\caption{Visualization of a sample image, its SLIC 
SP regions and \rbgraph graph from \coco dataset. In this figure, the extracted SP are $<50$ for better visualization. For the actual graphs with a maximum of $500$ SP and nodes in \pascal and \coco, refer to Appendix \ref{sec:viz_final_sp}.}
\label{fig:fig_viz_sp}
\vspace{-8pt}
\end{figure}

\textbf{Construction.}
After the superpixels extraction, we prepare the \rbgraph graph as illustrated step-wise in Figure \ref{fig:fig_viz_sp}.
Two superpixels nodes are connected with an edge if the node regions share a common boundary. We use the \texttt{rag\_boundary} functionality from \texttt{skimage} \cite{boulogne2014scikit} to extract the region boundaries. By construction, the dataset in this \rbgraph graph format has nodes with variying number of neighbors. This construction format also makes the graph more sparse with an average node degree of 5.6 for graphs averaging 479.40 node sizes. The initial feature of each superpixel node is 14 dimensional, 12-dim RGB feature value (mean, std, max, min) and 2-dim coordinates of the center of mass of pixel locations, and that of an edge between two nodes is a 2 dimensional vector where the first value is `weight' denoting the average of the Sobel filter \cite{engel2004real} pixel values along the boundary between the 2 adjacent regions, and the second value is `count' denoting the count of all pixels along this boundary.

\textbf{Performance Metric.} The performance metric is the macro weighted F1 score for the predicted node label and the ground truth node label.

\subsection{\coco}
Similar to \pascal, \coco is a node classification dataset based on the MS COCO image dataset \cite{lin2014microsoft} where each superpixel node denotes an image region belonging to a particular class. The original MS COCO image dataset is available under CC BY 4.0 License. We follow the same steps as in Section \ref{sec:pascalvoc} for the preparation of superpixels and the graphs in \coco for the \rbgraph graph format. Additional optional variants of the \coco datasets are included in Appendix \ref{app:dataset_details}.

\textbf{Statistics.} 
There are 123,286 graphs with a total of 58.7 million nodes in \coco where each graph corresponds to an image in MS COCO dataset \cite{lin2014microsoft}. The graphs prepared after the superpixels extraction have on average 476.88 nodes with complete statistics reported in Table \ref{tab:data_summary}.

\textbf{Task.} 
The learning task in \coco is node classification to predict a semantic segmentation label for each superpixel node out of 81 classes. We label each superpixel node with the same class label of the original pixel ground truth which is on the mean coordinates of the superpixel region.

\textbf{Splitting.}
In the MS COCO image dataset there are only train and validation sets available that we can use. In \coco, we maintain the original validation set as the new test set, while we sample 5,000 images from the original training set to generate the new validation set. Finally, there are 113,286 graphs, 5,000 graphs and 5,000 graphs in the resultant training, validation and test set, respectively.

\textbf{Performance Metric.} Similar to \pascal, the performance metric is the macro weighted F1 score for the predicted node label and the groundtruth node label.

\subsection{\pcqmcontact}
Molecular property prediction is one of the most popular tasks for benchmarking GNNs.
The usual task is to predict a biochemical property \cite{hu2020open, dwivedi2020benchmarking}.
Molecular datasets are very interesting for the study of Graph Transformers since their properties do not only depend on local graph structure defined by covalent bonds, but \textit{inter alia} also on long-range interactions that define the 3D folding of the molecules, their surface area, or the energy of their electronic orbitals \cite{ying2021graphormer,liu2022pretraining,stark2022-3d}.

However, existing benchmarks do not necessarily depend on long-range interactions and are noisy to properly evaluate Graph Transformers. For instance, the task in ZINC dataset from Dwivedi et al. \cite{dwivedi2020benchmarking} 
depends on a linear combination of local structures \cite{jin_junction_2018}. Thus, it is unclear whether there is any benefit in using a fully-connected Transformer instead of a standard message passing network \cite{rampasek2022GPS}. OGB \cite{hu2020open} offers a variety of datasets from biological assays, but they are often small, noisy, and it is unclear whether they would benefit from the long-range interactions of a Transformer. In \pcqmcontact, we design a task that explicitly requires LRI since it needs to understand the interaction between distant atoms.

\textbf{Statistics.} There are 529,434 graphs with a total of 15 million nodes in \pcqmcontact where each graph corresponds to a molecular graph with explicit hydrogens, and more details in Table \ref{tab:data_summary}. All graphs were taken from the PCQM4M training set with available 3D structure \cite{hu2021ogb_lsc} and filtered to only keep those with at least one contact.

\textbf{Task.} 
The task is to predict pairs of distant nodes (more than 5 hops away from each other in a molecule graph) that will be contacting with each other in the 3D space, i.e., the 3D distance between atoms will be smaller than 3.5{\AA}. The threshold of 3.5{\AA} is chosen to account for hydrogen bonds, one of the most common non-covalent interaction with a typical distance of 2.7 to 3.3 {\AA}
\cite{MCREE199991}. The 5-hop distance is chosen to avoid \textit{trivial} predictions between atoms that are close in the molecular graph and force the network to learn properties related to the 3D structure. Note that, contrarily to most benchmarks that represent hydrogens implicitly with node features, \pcqmcontact makes the hydrogens atoms explicit by adding a node.

Contact map prediction is therefore framed as inductive link prediction. 
That is, training molecules only have true positive links without hard negative labels whereas at validation and test time we predict contact links over new, unseen molecules.
The molecules are treated as relational graphs with learnable edge types (standard BondEncoder in OGB~\cite{hu2021ogb_lsc} e.g., single bond, double bond, triple bond), but the predictable contact link does not have an explicit edge type, so a link prediction decoder is a function $f(h, t)$ of probed head and tail nodes.

\textbf{Splitting.} We randomly split the dataset into 90\% (476,490 molecules) training split, 5\% (26,472 molecules) validation split, and 5\% (26,472 molecules) testing split.

\textbf{Performance Metric.}
In the absence of true negatives, we resort to the ranking metrics common in the knowledge graph link prediction literature~\cite{transe}. Given a query $(h, ?)$, we compute a scalar score for each other node in a graph as a tail $(h, t_i)$, and look for a rank of a true positive link. The true link has rank 1 if its score is the highest among all other links. We use standard ranking metrics Hits@1, Hits@3, and Mean Reciprocal Rank (MRR, \emph{aka} Inverse Harmonic Mean Rank~\cite{hoyt2022metrics}) in the \emph{filtered} setting~\cite{transe}. 
That is, if there exist several true links sharing the same head (or tail), i.e., $(h, t_1), (h, t_2), \dots, (h, t_k)$, evaluating each link separately, we filter out (mask) scores of other true tails setting their scores to $-\infty$ such that they do not interfere with the ranking procedure.

\subsection{Peptides molecular graphs}
\label{sec:peptides}

\begin{wrapfigure}{r}{0.35\textwidth}
    \centering
    \includegraphics[trim=30 5 900 30, clip, width=1.1\linewidth]{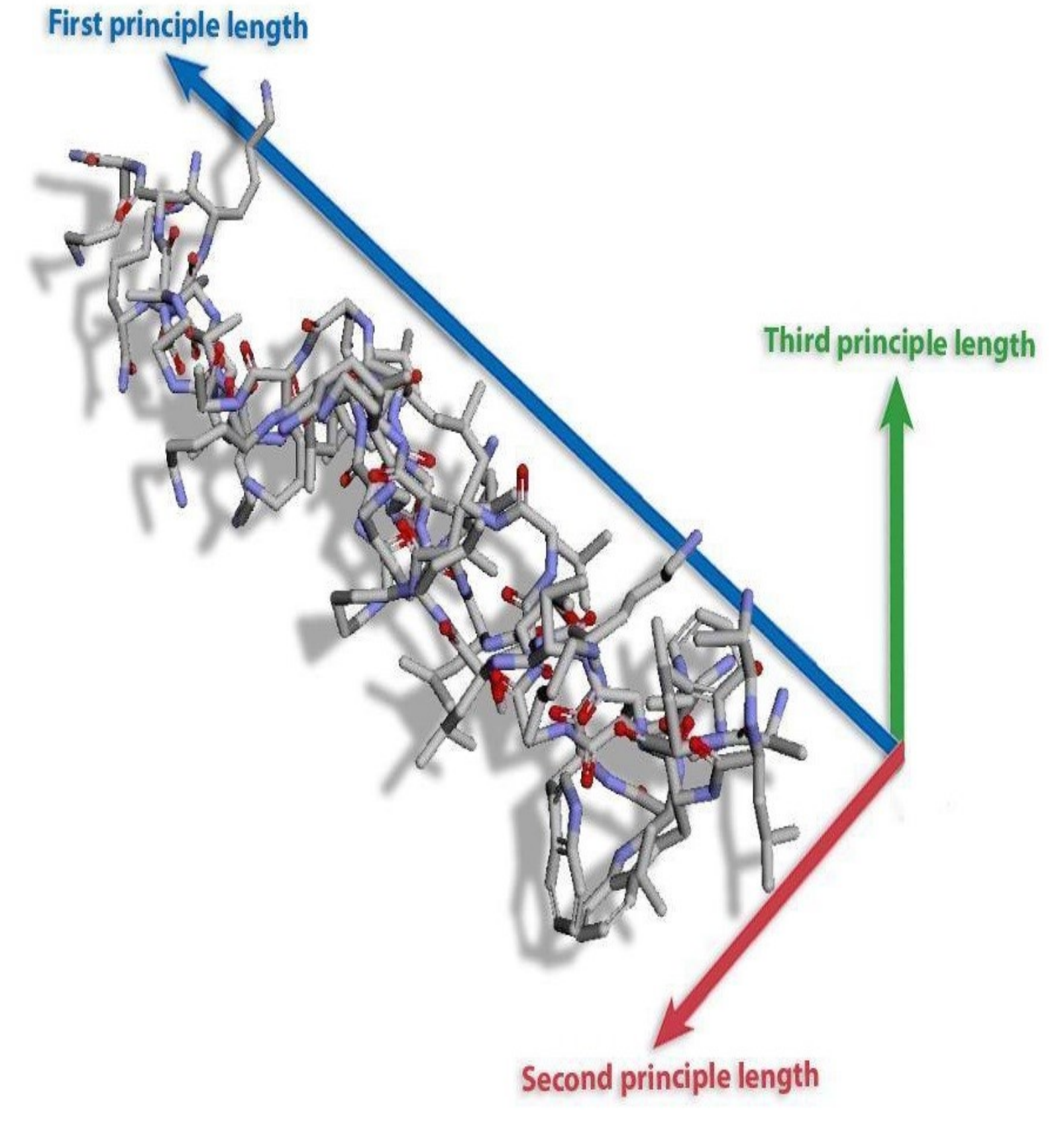}
    \includegraphics[trim=840 5 30 50, clip, width=1.1\linewidth]{images/peptide-3d-1.pdf}
    \vspace{-10pt}
    \caption{\textbf{Top:} 3D Visualization of "GLLGPLLKIAAKVGKNLL" peptide. \textbf{Bottom:} The molecular graph for the same peptide. }
    \label{fig:pep_3d_plot}
    \vspace*{-20pt}
\end{wrapfigure}

Peptides are short chains of amino acids that are abundant in nature as they serve many important biological functions \cite{singh2016satpdb}, but they are much shorter than proteins \cite{naturePeptideLearn}. Since each amino acid is composed of many heavy atoms, the molecular graph of a peptide is much larger than that of a small drug-like molecule. Peptides have about 6 times large diameter and 5 times more atoms than the \pcqmcontact dataset, but similar avg. degree of \textasciitilde 2. See Figure \ref{fig:pep_3d_plot} for an illustration. This makes them ideal for testing long-range dependencies in GNNs while still being able to fit an entire mini-batch on a single GPU.

Here we propose \pepfunc and \pepstruct datasets, derived from 15,535 peptides retrieved from SATPdb~\cite{singh2016satpdb}. Both datasets use the same set of graphs but differ in their prediction tasks.

\textbf{Construction.}
The graphs are derived such that the nodes correspond to the heavy (non-hydrogen) atoms of the peptides while the edges represent the bonds between them. We reuse the OGB molecular featurization \cite{hu2020open} that computes rich node and edge features from molecular SMILES.

In both \pepfunc and \pepstruct, recognizing local structures is very important for the model to even identify the original amino acids. Further, we do not include any 2D or 3D peptide structure information. The graphs correspond to 1D amino acids chains, which means it is important for the model to identify the location of an amino acid in the graph. Finally, with the peptides chains having different lengths and a strong variability in their graph diameters, any used graph positional or structural encoding needs to generalize well across various sizes and be computationally efficient.

\textbf{Statistics.} Both \pepfunc and \pepstruct consist of 15,535 graphs with a total of 2.3 million nodes, 
Table \ref{tab:data_summary}. All peptides were obtained from the SATPdb \cite{singh2016satpdb} database (an aggregate of multiple public-domain sources) that includes the sequence, molecular graph, function, and 3D structure of the peptides.

Previously introduced ENZYMES and PROTEINS datasets \cite{borgwardt2005protein}, that use the 3D structure of the folded proteins to build a graph of amino acids, are notably different from those that we propose here. In addition to more complex prediction tasks, our datasets are also larger in multiple ways. First, we derive 15,535 graphs, compared to theirs 600 and 1,113, respectively.
Second, we use heavy atoms as nodes and not the amino acids, resulting in larger graphs: on average 150.94 nodes per graph, compared to theirs 32.63 and 39.06, respectively. In terms of graph diameter, our graphs average 56.99 compared to theirs 10.92 and 11.62, respectively. Thus, the proposed \pepfunc and \pepstruct are better suited to benchmarking of graph Transformers or other expressive GNNs, as they contain larger graphs, more data points, and challenging tasks.

\subsubsection{\pepfunc}

\textbf{Task.}
\pepfunc is a multi-label graph classification dataset. There is a total of 10 classes based on the peptide function, e.g., \textit{Antibacterial, Antiviral, cell-cell communication}, and others. We treat it as a multi-label classification as a peptide can belong to several classes simultaneously; on average to 1.65 of the 10 classes. The labels are imbalanced, only 16.5\% of the data is in the positive class, with the richest class having 62.7\% positives and the poorest 1.9\%. The correlation between individual classes is shown in Figure~\ref{fig:pep_clas_corr}.

\textbf{Splitting.}
For the purpose of split computation, the data is first aggregated into meta-classes by considering the concatenation of all 10 original labels of a data point as its meta-class. Meta-classes with less than 10 occurrences are pooled into one meta-class. Then we apply stratified splitting to generate balanced train--valid--test dataset splits; we use the ratio of 70\%--15\%--15\%, respectively.

\textbf{Performance Metric.}
We choose the unweighted mean Average Precision (AP). This metric measures the area under the \textit{precision-recall curve}, and is also used for \texttt{ogbg-molpcba} in OGB, a dataset with similar imbalanced multi-label classification.

\subsubsection{\pepstruct}

\textbf{Task.}
\pepstruct is a multi-label graph regression dataset based on the 3D structure of the peptides. It consists of the same graphs as \pepfunc, but with different task. Here we aim to predict aggregated 3D properties of the peptides at the graph level. 
The properties (normalized to zero mean and unit standard deviation) include: \textit{Inertia\_mass}, \textit{Inertia\_valence}, \textit{Length}, \textit{sphericity}, and \textit{plane\_best\_fit}. How we derive these properties is described in Appendix~\ref{app:peptstruct_description} and their correlations are shown in Figure~\ref{fig:pep_clas_corr}.
These new tasks are expected to directly benefit from the full-connectivity of a Transformer, because they require implicit understanding of complex 3D interactions. To adapt these tasks to the general graph learning setting, we avoided the prediction of pairwise node distances, which would require specialized methods from the conformer generation literature \cite{ganea2021geomol, LearningGradFields, LearningGenerativeDyn_MinkaiXu}.

\textbf{Splitting.}
The data splits are identical to \pepfunc. Since structure is related to functionality, this also ensures that the structures in the testing set represent well the training and validation sets.

\textbf{Performance Metric.}
For the \pepstruct dataset, we use the Mean Absolute Error (MAE), typically selected for molecular property regression datasets. We further track the Coefficient of Determination ($\mathrm{R}^2$) on each task, and compute its unweighted mean across tasks.

\section{Experiments and Discussion}

\subsection{Baseline experiments\textsuperscript{4}}\footnotetext[4]{For a reassessment of the original baselines on all the datasets as well as extensions of \pcqmcontact's metric, we refer to the paper \cite{tonshoff2023did} by Tönshoff et al., 2023. In addition, we direct the readers to \url{https://paperswithcode.com/dataset/pascalvoc-sp} for an up-to-date leaderboard of the benchmark with model contributions from the community.}
We conduct baseline experiments on our proposed LRGB datasets by training and evaluating two GNN classes: (i) local MP-GNNs, and (ii) fully connected Graph Transformers. We adopt fair and rigorous experimental settings for all our experiments in order to present reliable comparison between the two GNN classes. The former models do not directly include any mechanism to model LRI, while the latter are by design fully connected and can propagate long-range signals, which are required for the proposed benchmarks. For the baselines, we select GCN~\cite{kipf2016gcn}, GCNII \cite{chen2020GCNII}, GINE~\cite{xu2018powerful_gin, hu2019strategies} and GatedGCN~\cite{bresson2017gatedGCN} models from the local MP-GNN class, and fully connected Transformer \cite{vaswani_2017_attention} with Laplacian PE (LapPE) \cite{dwivedi2020benchmarking, dwivedi2020generalization} and SAN \cite{kreuzer2021rethinking} models from the Transformer class. In order to facilitate fair comparison and reliable discussion of the observed trends, we choose hyperparameters of the aforementioned baselines while keeping to a budget of 500k learnable parameters. Detailed experimental setup and hyperparameters are provided in Appendix~\ref{app:exp_details}.

\begin{table}[t]
    \caption{Baseline experiments for \pascal and \coco with \rbgraph graph on SLIC compactness 30 for node classification task (Extended results for all graph formats in Table \ref{tab:experiments_superpixels_full}). Performance metric is macro F1 on the respective splits (Higher is better). All experiments are run 4 times with 4 different seeds. 
    The MP-GNN models are 8 layers deep, while the transformer-based models have 4 layers in order to maintain comparable hidden representation size at the fixed parameter budget of 500k. *The SAN model under-fitted the \coco dataset since it required more budget than the 60 hours allowed on A100 GPUs. \textbf{Bold}: Best score.
    }
    \label{tab:experiments_superpixels}
    \begin{adjustwidth}{-2.5 cm}{-2.5 cm}\centering
    \scalebox{0.87}{
    \setlength\tabcolsep{4pt} 
    \begin{tabular}{l c A B r A B}
    \toprule
    \multirow{2}{*}{\textbf{Model}} & \multirow{2}{*}{\hspace*{-2em}\textbf{\# Params}} & \multicolumn{2}{c}{\pascal} & \multirow{2}{*}{\textbf{\# Params}} & \multicolumn{2}{c}{\coco} \\ \cmidrule(lr){3-4}\cmidrule(lr){6-7}
    & & \textbf{Train F1} & \textbf{Test F1 $\uparrow$} & & \textbf{Train F1} & \textbf{Test F1 $\uparrow$}\\
    \midrule 
    GCN &496k &0.1450$\pm$0.0125 &0.1268$\pm$0.0060 & 509k &0.0948$\pm$0.0014 &0.0841$\pm$0.0010 \\
    GCNII & 492k & 0.2272$\pm$0.0245 & 0.1698$\pm$0.0080 & 505k & 0.2020$\pm$0.0127 & 0.1404$\pm$0.0011\\
    GINE & 505k &0.2088$\pm$0.0268 &0.1265$\pm$0.0076 &515k &0.2100$\pm$0.0041 &0.1339$\pm$0.0044 \\
    GatedGCN & 502k &0.3552$\pm$0.0451 &  0.2873$\pm$0.0219 & 509k &0.3167$\pm$0.0059 &\first{0.2641$\pm$0.0045} \\
    GatedGCN+LapPE &502k &0.3512$\pm$0.0167 &0.2860$\pm$0.0085 &509k &0.3102$\pm$0.0112 &0.2574$\pm$0.0034\\ \cmidrule(l){1-7}
    Transformer+LapPE &501k &0.7170$\pm$0.0048 &0.2694$\pm$0.0098 &508k &0.3912$\pm$0.0098 &\second{0.2618$\pm$0.0031} \\
    SAN+LapPE &531k &0.5723$\pm$0.0427 &\first{0.3230$\pm$0.0039} &536k &0.2830$\pm$0.0246* &0.2592$\pm$0.0158* \\
    SAN+RWSE &468k &0.5819$\pm$0.0331 &\second{0.3216$\pm$0.0027} &474k &0.2657$\pm$0.0224* &0.2434$\pm$0.0156* \\
    \bottomrule
    \end{tabular}
    }
    \end{adjustwidth}
\end{table}

\subsection{Results and Analysis}
The baseline results for \pascal and \coco benchmarks are reported in Table \ref{tab:experiments_superpixels}, for \pepfunc and \pepstruct in Table \ref{tab:experiments_peptides}, and for \pcqmcontact in Table \ref{tab:experiments_pcqmcontact}. We also provide additional baseline results for all MP-GNNs with fewer layers ($L=2$) in Appendix \ref{app:2layer-mpnns}. Our aim is to address the following main questions through the analysis of these results:
\vspace{-3pt}
\begin{enumerate}[label=(\roman*), leftmargin=2.5em]
    \item Is a local feature aggregation, modeled using MP-GNNs with fewer layers, enough for the proposed tasks in LRGB?
    \item Do we observe a visible separation in learning and generalization of models with enhanced capability to capture LRIs when compared against local MP-GNNs on the proposed benchmark?
    \item Does the use of positional encodings, that contribute critical structural information,
    improve MP-GNN performance on the proposed datasets?
    \item What are the challenges and future discoveries that can be facilitated by the new benchmarks? 
\end{enumerate}

\vspace{8pt}

\begin{table}[t]
    \caption{Baselines for \pepfunc (graph classification) and \pepstruct (graph regression). Performance metric is Average Precision (AP) for classification and MAE for regression (see Table~\ref{tab:pep_r2} for extended results with $\mathrm{R}^2$ metric). Each experiment was run with 4 different seeds. All MP-GNN models have 5 layers, while the Transformer-based models have 4 layers. \textbf{Bold}: Best score.
    }
    \label{tab:experiments_peptides}
    \begin{adjustwidth}{-2.5 cm}{-2.5 cm}\centering
    \scalebox{0.9}{
    \setlength\tabcolsep{4pt} 
    \begin{tabular}{l c A B A C}\toprule
    \multirow{2}{*}{\textbf{Model}} &\multirow{2}{*}{\textbf{\# Params.}} &\multicolumn{2}{c}{\pepfunc} &\multicolumn{2}{c}{\pepstruct} \\ \cmidrule(lr){3-4} \cmidrule(lr){5-6}
    & &\textbf{Train AP} &\textbf{Test AP $\uparrow$} &\textbf{Train MAE} &\textbf{Test MAE $\downarrow$} \\\midrule
    GCN &508k &0.8840$\pm$0.0131 &0.5930$\pm$0.0023 &0.2939$\pm$0.0055 &0.3496$\pm$0.0013 \\
    GCNII & 505k & 0.7271$\pm$0.0278 & 0.5543$\pm$0.0078 & 0.2957$\pm$0.0025 & 0.3471$\pm$0.0010\\
    GINE &476k &0.7682$\pm$0.0154 &0.5498$\pm$0.0079 &0.3116$\pm$0.0047 &0.3547$\pm$0.0045 \\
    GatedGCN &509k &0.8695$\pm$0.0402 &0.5864$\pm$0.0077 &0.2761$\pm$0.0032 &0.3420$\pm$0.0013 \\
    GatedGCN+RWSE &506k &0.9131$\pm$0.0321 &0.6069$\pm$0.0035 &0.2578$\pm$0.0116 &0.3357$\pm$0.0006 \\ \midrule
    Transformer+LapPE &488k &0.8438$\pm$0.0263 &0.6326$\pm$0.0126 &0.2403$\pm$0.0066 &\first{0.2529$\pm$0.0016} \\
    SAN+LapPE &493k &0.8217$\pm$0.0280 &\second{0.6384$\pm$0.0121} &0.2822$\pm$0.0108 &0.2683$\pm$0.0043 \\
    SAN+RWSE &500k &0.8612$\pm$0.0219 &\first{0.6439$\pm$0.0075} &0.2680$\pm$0.0038 &\second{0.2545$\pm$0.0012} \\
    \bottomrule
    \end{tabular}
    }
    \vspace{-1pt}
    \end{adjustwidth}
\end{table}

\begin{table}[t]
    \caption{Baseline performance on \pcqmcontact (link prediction). Each experiment was repeated with 4 different random seeds. The evaluated models have 5 (MP-GNN models) or 4 (Transformer-based models) layers with approximately 500k learnable parameters. \textbf{Bold}: Best score.
    }
    \label{tab:experiments_pcqmcontact}
    \begin{adjustwidth}{-2.5 cm}{-2.5 cm}\centering
    \scalebox{0.9}{
    \setlength\tabcolsep{3pt} 
    \begin{tabular}{l c B B B B}\toprule
    \textbf{Model} &\textbf{\# Params.} &\textbf{Test Hits@1 $\uparrow$} &\textbf{Test Hits@3 $\uparrow$} &\textbf{Test Hits@10 $\uparrow$} &\textbf{Test MRR $\uparrow$} \\\midrule
    GCN &504k &0.1321$\pm$0.0007 &0.3791$\pm$0.0004 &0.8256$\pm$0.0006 &0.3234$\pm$0.0006 \\
    GCNII & 501k & 0.1325$\pm$0.0009 & 0.3607$\pm$0.0003 & 0.8116$\pm$0.0009 & 0.3161$\pm$0.0004\\
    GINE &517k &0.1337$\pm$0.0013 &0.3642$\pm$0.0043 &0.8147$\pm$0.0062 &0.3180$\pm$0.0027 \\
    GatedGCN &527k &0.1279$\pm$0.0018 &0.3783$\pm$0.0004 &0.8433$\pm$0.0011 &0.3218$\pm$0.0011 \\
    GatedGCN+RWSE &524k &0.1288$\pm$0.0013 &0.3808$\pm$0.0006 &0.8517$\pm$0.0005 &0.3242$\pm$0.0008 \\ \midrule
    Transformer+LapPE &502k &0.1221$\pm$0.0011 &0.3679$\pm$0.0033 &0.8517$\pm$0.0039 &0.3174$\pm$0.0020 \\
    SAN+LapPE &499k &\first{0.1355$\pm$0.0017} &0.4004$\pm$0.0021 &0.8478$\pm$0.0044 &\first{0.3350$\pm$0.0003} \\
    SAN+RWSE &509k &0.1312$\pm$0.0016 &\first{0.4030$\pm$0.0008} &\first{0.8550$\pm$0.0024} &\second{0.3341$\pm$0.0006} \\
    \bottomrule
    \end{tabular}
    }
    \end{adjustwidth}
\end{table}

\textbf{Simple instances of local MP-GNNs perform poorly on the proposed LRGB datasets.} As shown by the results in Tables \ref{tab:experiments_superpixels}, \ref{tab:experiments_peptides} and \ref{tab:experiments_pcqmcontact}, GCN and GINE, which depend on local feature aggregation from node neighborhoods using simple aggregation functions, perform poorly on all datasets except \pepfunc. This is consistent with the empirical findings in \cite{alon2020bottleneck} where GCN and GIN suffer from over-squashing to a greater extent than GAT, an attention based MP-GNN \cite{velikovic2017gat}.

\textbf{Shallow MP-GNNs that gather information from only close neighbors underfit.} The comparison of shallow MP-GNN baselines ($L=2$, see Appendix \ref{app:2layer-mpnns}) with deeper ones ($L=5,8$, see Tables \ref{tab:experiments_superpixels}-\ref{tab:experiments_pcqmcontact}) shows, that models with information aggregation limited to only a few hops significantly underfit and provide poor generalization on test set, as compared to MP-GNNs with the increased receptive field.
This points towards the proposed benchmarks being \textit{different} from several classical benchmarks such as Cora or Citeseer, \textit{inter alia} \cite{chen2020GCNII}, where shallow GCNs \cite{kipf2016gcn} fared better than deeper GCNs. They are thus more suitable for evaluating GNNs with deeper architectures, increased receptive fields, as well as long-range modeling, for which we provide a further study in the following analysis.

\textbf{Transformers operating on fully-connected graph show the best performance.} It can be observed that the Transformer model and the SAN, which is an improved Transformer, rank among the best performing baselines in Tables~\ref{tab:experiments_superpixels},~\ref{tab:experiments_peptides}~and~\ref{tab:experiments_pcqmcontact}, except for GatedGCN vs. Transformer in Table \ref{tab:experiments_superpixels}. The gap in performance seems the most distinct for
\pepfunc and \pepstruct 
among all datasets. This can be attributed to the long-range design of the task, the contribution of non-local information, and the graph size statistics of these datasets, as discussed in Section \ref{sec:peptides}. On \pascal and \pcqmcontact, SAN performs comparatively better than vanilla Transformer+LapPE, suggesting that full connections in graphs should be used in non-trivial manner for LRI enabled models to do well. 
Finally, on \coco the observed difference between Transformer and GatedGCN is minor, while SAN did not manage to converge within a 60h computational limit. This exposes the scalability drawbacks of current Transformer-based models, that would likely benefit from increased parameter and computational budget, see Table~\ref{tab:run-times}. In Appendix \ref{app:attn-analysis}, we show additional investigation, with visualizations, on how Transformer exhibits attention patterns beyond local neighborhoods in general.

\textbf{Discussion on LRI characterizing factors.} 
To a major extent, the reasons behind the fully-connected Transformer baselines being able to excel in the proposed LRGB datasets can be linked to one or all of the characterizing factors that were discussed in Section \ref{sec:characterizing}. For instance, the nature of task of \texttt{Peptides-*} datasets, along with their substantial graph statistics (\textit{i.e.}, avg. nodes, avg. shortest paths and avg. diameter as reported in Table \ref{tab:data_summary}) can help explain how long range dependencies are a must to do well on such tasks. Similarly, for \pcqmcontact, even if the graph sizes are small, the task of predicting pairs of distant nodes makes it a suitable LRGB dataset as shown by, e.g., Test Hits@3 scores of SAN against the local MP-GNNs in Table \ref{tab:experiments_pcqmcontact}.

\textbf{Challenges and future directions.}
\underline{First}, the use of positional encoding alone contributes to little or no gain in performance on the proposed datasets. See the scores of GatedGCN augmented with LapPE or RWSE in Tables \ref{tab:experiments_superpixels}, \ref{tab:experiments_peptides} and \ref{tab:experiments_pcqmcontact} to this end. We hope such results to influence further exploration of powerful approaches to incorporate global structural and positional encoding in LRI enabled models, where the proposed LRGB can be used to conveniently evaluate the novel approaches. 
\underline{Second}, the scores against each performance metrics in Tables \ref{tab:experiments_superpixels}-\ref{tab:experiments_pcqmcontact} exhibit the current limitations of Transformers for graph learning and suggest that there is still a large window to fulfil by the better design of Graph Transformers that can make use of irregular sparse structure information, as well as propagate long range interactions. 
\underline{Finally}, it must be noted that as we proceed towards evaluating Graph Transformers on long range benchmarks, such as our proposed LRGB with up to 479.40 avg. nodes, 58.79 million total nodes and 332.09 million total edges in a dataset, trivial $O(N^2)$ Transformers may be computationally inefficient to scale. To this end, research is also imperative on efficient or linear Transformers for graphs.

\section{Conclusion}
In this paper, we present the Long Range Graph Benchmark (LRGB) consisting of 5 datasets for node, edge and graph-level prediction tasks. Through our study of multiple characterizing factors, we argue that the proposed datasets' size and tasks makes these ideal to evaluate and develop models enabled with long-range dependencies. This is empirically verified with extensive baseline experiments using both local and non-local GNN classes showing that Transformers significantly outperform message passing on the proposed datasets.
The increasing interest in the development of Transformers for graph representation learning raised the need for the creation of a dedicated LRGB and we fulfil this gap through our work. We believe our proposed benchmark can be leveraged to prototype new ideas and provide an accurate ranking of a model's capturing of LRIs.

\begin{ack}
This work was partially funded by IVADO (Institut de valorisation des données) grant PRF-2019-3583139727 and Canada CIFAR AI Chair~[\emph{G.W.}]. Ali Parviz is supported by NSF
Award \#2039863. This research is supported by Nanyang Technological University, under SUG Grant (020724-00001). The content provided here is solely the responsibility of the authors and does not necessarily represent the official views of the funding agencies.
\end{ack}

\bibliography{main}

\begin{thebibliography}{10}

\bibitem{achanta2010slic}
Radhakrishna Achanta, Appu Shaji, Kevin Smith, Aurelien Lucchi, Pascal Fua, and Sabine Süsstrunk.
\newblock {SLIC Superpixels} compared to state-of-the-art superpixel methods.
\newblock {\em IEEE Transactions on Pattern Analysis and Machine Intelligence}, 34(11):2274--2282, 2012.

\bibitem{alon2020bottleneck}
Uri Alon and Eran Yahav.
\newblock On the bottleneck of graph neural networks and its practical implications.
\newblock In {\em International Conference on Learning Representations}, 2021.

\bibitem{bodnar2021weisfeiler}
Cristian Bodnar, Fabrizio Frasca, Nina Otter, Yu~Guang Wang, Pietro Li{\`o}, Guido~F Montufar, and Michael Bronstein.
\newblock Weisfeiler and lehman go cellular: {CW} networks.
\newblock {\em Advances in Neural Information Processing Systems}, 34, 2021.

\bibitem{transe}
Antoine Bordes, Nicolas Usunier, Alberto Garc{\'{\i}}a{-}Dur{\'{a}}n, Jason Weston, and Oksana Yakhnenko.
\newblock Translating embeddings for modeling multi-relational data.
\newblock In {\em Advances in Neural Information Processing Systems}, pages 2787--2795, 2013.

\bibitem{borgwardt2005protein}
Karsten~M Borgwardt, Cheng~Soon Ong, Stefan Sch{\"o}nauer, SVN Vishwanathan, Alex~J Smola, and Hans-Peter Kriegel.
\newblock Protein function prediction via graph kernels.
\newblock {\em Bioinformatics}, 21:i47--i56, 2005.

\bibitem{boulogne2014scikit}
Fran{\c{c}}ois Boulogne, Joshua~D Warner, and Emmanuelle Neil~Yager.
\newblock Scikit-image: Image processing in python.
\newblock {\em J. PeerJ}, 2:453, 2014.

\bibitem{bouritsas2022improving}
Giorgos Bouritsas, Fabrizio Frasca, Stefanos~P Zafeiriou, and Michael Bronstein.
\newblock Improving graph neural network expressivity via subgraph isomorphism counting.
\newblock {\em IEEE Transactions on Pattern Analysis and Machine Intelligence}, 2022.

\bibitem{bresson2017gatedGCN}
Xavier Bresson and Thomas Laurent.
\newblock Residual gated graph convnets.
\newblock {\em arXiv:1711.07553}, 2017.

\bibitem{bronstein2021geometric}
Michael~M Bronstein, Joan Bruna, Taco Cohen, and Petar Veli{\v{c}}kovi{\'c}.
\newblock Geometric deep learning: Grids, groups, graphs, geodesics, and gauges.
\newblock {\em arXiv:2104.13478}, 2021.

\bibitem{chen2017stochastic}
Jianfei Chen, Jun Zhu, and Le~Song.
\newblock Stochastic training of graph convolutional networks with variance reduction.
\newblock In {\em International Conference on Machine Learning}, 2018.

\bibitem{chen2020GCNII}
Ming Chen, Zhewei Wei, Zengfeng Huang, Bolin Ding, and Yaliang Li.
\newblock Simple and deep graph convolutional networks.
\newblock In {\em International Conference on Machine Learning}, 2020.

\bibitem{defferrard2016convolutional}
Micha{\"e}l Defferrard, Xavier Bresson, and Pierre Vandergheynst.
\newblock Convolutional neural networks on graphs with fast localized spectral filtering.
\newblock {\em Advances in neural information processing systems}, 29, 2016.

\bibitem{dwivedi2020generalization}
Vijay~Prakash Dwivedi and Xavier Bresson.
\newblock A generalization of transformer networks to graphs.
\newblock {\em AAAI Workshop on Deep Learning on Graphs: Methods and Applications}, 2021.

\bibitem{dwivedi2020benchmarking}
Vijay~Prakash Dwivedi, Chaitanya~K Joshi, Anh~Tuan Luu, Thomas Laurent, Yoshua Bengio, and Xavier Bresson.
\newblock Benchmarking graph neural networks.
\newblock {\em arXiv:2003.00982}, 2020.

\bibitem{dwivedi2022graph}
Vijay~Prakash Dwivedi, Anh~Tuan Luu, Thomas Laurent, Yoshua Bengio, and Xavier Bresson.
\newblock Graph neural networks with learnable structural and positional representations.
\newblock In {\em International Conference on Learning Representations}, 2022.

\bibitem{dwivedi_vijay_prakash_2022_6975830}
Vijay~Prakash Dwivedi, Ladislav Rampášek, Mikhail Galkin, Ali Parviz, Guy Wolf, Anh~Tuan Luu, and Dominique Beaini.
\newblock {Zenodo repository of LRGB: Long Range Graph Benchmark}.
\newblock \url{https://doi.org/10.5281/zenodo.6975830}, 2022.

\bibitem{engel2004real}
Klaus Engel, Markus Hadwiger, Joe~M Kniss, Aaron~E Lefohn, Christof~Rezk Salama, and Daniel Weiskopf.
\newblock Real-time volume graphics.
\newblock In {\em ACM Siggraph 2004 Course Notes}. CRC Press, 2004.

\bibitem{everingham2010pascal}
Mark Everingham, Luc Van~Gool, Christopher~KI Williams, John Winn, and Andrew Zisserman.
\newblock {The PASCAL visual object classes (VOC) challenge}.
\newblock {\em International journal of computer vision}, 88(2):303--338, 2010.

\bibitem{FeyLenssen2019PyG}
Matthias Fey and Jan~Eric Lenssen.
\newblock Fast graph representation learning with {PyTorch Geometric}.
\newblock In {\em ICLR Workshop on Representation Learning on Graphs and Manifolds}, 2019.

\bibitem{freitas2020large}
Scott Freitas, Yuxiao Dong, Joshua Neil, and Duen~Horng Chau.
\newblock A large-scale database for graph representation learning.
\newblock In {\em 35th Conference on Neural Information Processing Systems: Datasets and Benchmarks Track}, 2021.

\bibitem{ganea2021geomol}
Octavian-Eugen Ganea, Lagnajit Pattanaik, Connor~W. Coley, Regina Barzilay, Klavs~F. Jensen, William~H. Green, and Tommi~S. Jaakkola.
\newblock {GeoMol}: Torsional geometric generation of molecular {3D} conformer ensembles.
\newblock In {\em Advances in Neural Information Process Systems}, 2021.

\bibitem{gilmer2017neural_mpnn}
Justin Gilmer, Samuel~S Schoenholz, Patrick~F Riley, Oriol Vinyals, and George~E Dahl.
\newblock Neural message passing for quantum chemistry.
\newblock In {\em International conference on machine learning}, pages 1263--1272. PMLR, 2017.

\bibitem{gu2020implicit}
Fangda Gu, Heng Chang, Wenwu Zhu, Somayeh Sojoudi, and Laurent El~Ghaoui.
\newblock Implicit graph neural networks.
\newblock {\em Advances in Neural Information Processing Systems}, 33:11984--11995, 2020.

\bibitem{hamilton}
William~L. Hamilton.
\newblock Graph representation learning.
\newblock {\em Synthesis Lectures on Artificial Intelligence and Machine Learning}, 14(3):1--159, 2020.

\bibitem{hoyt2022metrics}
Charles~Tapley Hoyt, Max Berrendorf, Mikhail Gaklin, Volker Tresp, and Benjamin~M. Gyori.
\newblock {A Unified Framework for Rank-based Evaluation Metrics for Link Prediction in Knowledge Graphs}.
\newblock In {\em Graph Learning Benchmarks Workshop at The WebConf 2022}, 2022.

\bibitem{hu2021ogb_lsc}
Weihua Hu, Matthias Fey, Hongyu Ren, Maho Nakata, Yuxiao Dong, and Jure Leskovec.
\newblock {OGB-LSC}: A large-scale challenge for machine learning on graphs.
\newblock In {\em 35th Conference on Neural Information Processing Systems Datasets and Benchmarks Track (Round 2)}, 2021.

\bibitem{hu2020open}
Weihua Hu, Matthias Fey, Marinka Zitnik, Yuxiao Dong, Hongyu Ren, Bowen Liu, Michele Catasta, and Jure Leskovec.
\newblock Open graph benchmark: Datasets for machine learning on graphs.
\newblock In {\em Advances in neural information processing systems}, 2020.

\bibitem{hu2019strategies}
Weihua Hu, Bowen Liu, Joseph Gomes, Marinka Zitnik, Percy Liang, Vijay Pande, and Jure Leskovec.
\newblock Strategies for pre-training graph neural networks.
\newblock In {\em International Conference on Learning Representations}, 2020.

\bibitem{irwin2012zinc}
John~J Irwin, Teague Sterling, Michael~M Mysinger, Erin~S Bolstad, and Ryan~G Coleman.
\newblock {ZINC}: a free tool to discover chemistry for biology.
\newblock {\em Journal of chemical information and modeling}, 52(7):1757--1768, 2012.

\bibitem{jin_junction_2018}
Wengong Jin, Regina Barzilay, and Tommi Jaakkola.
\newblock Junction tree variational autoencoder for molecular graph generation.
\newblock In {\em International conference on machine learning}, pages 2323--2332. PMLR, 2018.

\bibitem{kim2021transformers}
Jinwoo Kim, Saeyoon Oh, and Seunghoon Hong.
\newblock Transformers generalize deepsets and can be extended to graphs \& hypergraphs.
\newblock {\em Advances in Neural Information Processing Systems}, 34:28016--28028, 2021.

\bibitem{kingma2017adam}
Diederik~P Kingma and Jimmy Ba.
\newblock Adam: A method for stochastic optimization.
\newblock In {\em International Conference on Learning Representations}, 2015.

\bibitem{kipf2016gcn}
Thomas~N Kipf and Max Welling.
\newblock Semi-supervised classification with graph convolutional networks.
\newblock In {\em International Conference on Learning Representations}, 2017.

\bibitem{kreuzer2021rethinking}
Devin Kreuzer, Dominique Beaini, Will Hamilton, Vincent L{\'e}tourneau, and Prudencio Tossou.
\newblock Rethinking graph transformers with spectral attention.
\newblock {\em Advances in Neural Information Processing Systems}, 34, 2021.

\bibitem{li_distance_2020}
Pan Li, Yanbang Wang, Hongwei Wang, and Jure Leskovec.
\newblock Distance encoding: Design provably more powerful neural networks for graph representation learning.
\newblock In {\em Advances in Neural Information Processing Systems}, 2020.

\bibitem{lim2021new}
Derek Lim, Xiuyu Li, Felix Hohne, and Ser-Nam Lim.
\newblock New benchmarks for learning on non-homophilous graphs.
\newblock {\em arXiv preprint arXiv:2104.01404}, 2021.

\bibitem{lim2022sign}
Derek Lim, Joshua Robinson, Lingxiao Zhao, Tess Smidt, Suvrit Sra, Haggai Maron, and Stefanie Jegelka.
\newblock Sign and basis invariant networks for spectral graph representation learning.
\newblock {\em arXiv:2202.13013}, 2022.

\bibitem{lin2014microsoft}
Tsung-Yi Lin, Michael Maire, Serge Belongie, James Hays, Pietro Perona, Deva Ramanan, Piotr Doll{\'a}r, and C~Lawrence Zitnick.
\newblock Microsoft {COCO}: Common objects in context.
\newblock In {\em European conference on computer vision}, pages 740--755. Springer, 2014.

\bibitem{liu2022mgnni}
Juncheng Liu, Bryan Hooi, Kenji Kawaguchi, and Xiaokui Xiao.
\newblock Mgnni: Multiscale graph neural networks with implicit layers.
\newblock {\em arXiv preprint arXiv:2210.08353}, 2022.

\bibitem{liu2021eignn}
Juncheng Liu, Kenji Kawaguchi, Bryan Hooi, Yiwei Wang, and Xiaokui Xiao.
\newblock Eignn: Efficient infinite-depth graph neural networks.
\newblock {\em Advances in Neural Information Processing Systems}, 34:18762--18773, 2021.

\bibitem{liu2022pretraining}
Shengchao Liu, Hanchen Wang, Weiyang Liu, Joan Lasenby, Hongyu Guo, and Jian Tang.
\newblock Pre-training molecular graph representation with 3d geometry.
\newblock In {\em International Conference on Learning Representations}, 2022.

\bibitem{ma2021deep}
Yao Ma and Jiliang Tang.
\newblock {\em Deep learning on graphs}.
\newblock Cambridge University Press, 2021.

\bibitem{MCREE199991}
Duncan~E McRee.
\newblock {\em Practical protein crystallography (2nd Edition)}.
\newblock Academic Press, 1999.

\bibitem{mialon2021graphit}
Gr{\'e}goire Mialon, Dexiong Chen, Margot Selosse, and Julien Mairal.
\newblock {GraphiT}: Encoding graph structure in transformers.
\newblock {\em arXiv:2106.05667}, 2021.

\bibitem{morris2020weisfeiler}
Christopher Morris, Gaurav Rattan, and Petra Mutzel.
\newblock {W}eisfeiler and {L}eman go sparse: Towards scalable higher-order graph embeddings.
\newblock {\em Advances in Neural Information Processing Systems}, 33:21824--21840, 2020.

\bibitem{morris2019weisfeiler}
Christopher Morris, Martin Ritzert, Matthias Fey, William~L Hamilton, Jan~Eric Lenssen, Gaurav Rattan, and Martin Grohe.
\newblock Weisfeiler and leman go neural: Higher-order graph neural networks.
\newblock In {\em Proceedings of the AAAI Conference on Artificial Intelligence}, volume~33, pages 4602--4609, 2019.

\bibitem{naturePeptideLearn}
Nature.com.
\newblock {L}earn {S}cience at {S}citable: Peptide.
\newblock \url{https://www.nature.com/scitable/definition/peptide-317/}.
\newblock [Accessed 08-Jun-2022].

\bibitem{pascalvoc}
PascalVOC.
\newblock Visual object classes challenge 2011 (voc2011).
\newblock \url{http://host.robots.ox.ac.uk/pascal/VOC/voc2011/index.html}.
\newblock [Accessed 08-Jun-2022].

\bibitem{pei2020geom}
Hongbin Pei, Bingzhe Wei, Kevin Chen-Chuan Chang, Yu~Lei, and Bo~Yang.
\newblock Geom-gcn: Geometric graph convolutional networks.
\newblock {\em arXiv preprint arXiv:2002.05287}, 2020.

\bibitem{rampavsek2021hierarchical}
Ladislav Ramp{\'a}{\v{s}}ek and Guy Wolf.
\newblock Hierarchical graph neural nets can capture long-range interactions.
\newblock In {\em 2021 IEEE 31st International Workshop on Machine Learning for Signal Processing (MLSP)}, pages 1--6. IEEE, 2021.

\bibitem{rampasek2022GPS}
Ladislav Ramp\'{a}\v{s}ek, Mikhail Galkin, Vijay~Prakash Dwivedi, Anh~Tuan Luu, Guy Wolf, and Dominique Beaini.
\newblock {Recipe for a General, Powerful, Scalable Graph Transformer}.
\newblock {\em Neural Information Processing Systems (NeurIPS)}, 2022.

\bibitem{LearningGradFields}
Chence Shi, Shitong Luo, Minkai Xu, and Jian Tang.
\newblock Learning gradient fields for molecular conformation generation.
\newblock In {\em International Conference on Machine Learning}, volume 139, pages 9558--9568. {PMLR}, 2021.

\bibitem{shi2022benchmarking}
Yu~Shi, Shuxin Zheng, Guolin Ke, Yifei Shen, Jiacheng You, Jiyan He, Shengjie Luo, Chang Liu, Di~He, and Tie-Yan Liu.
\newblock Benchmarking graphormer on large-scale molecular modeling datasets.
\newblock {\em arXiv:2203.04810}, 2022.

\bibitem{singh2016satpdb}
Sandeep Singh, Kumardeep Chaudhary, Sandeep~Kumar Dhanda, Sherry Bhalla, Salman~Sadullah Usmani, Ankur Gautam, Abhishek Tuknait, Piyush Agrawal, Deepika Mathur, and Gajendra~PS Raghava.
\newblock {SATPdb}: a database of structurally annotated therapeutic peptides.
\newblock {\em Nucleic acids research}, 44(D1):D1119--D1126, 2016.

\bibitem{stachenfeld2020graph}
Kimberly Stachenfeld, Jonathan Godwin, and Peter Battaglia.
\newblock Graph networks with spectral message passing.
\newblock {\em arXiv:2101.00079}, 2020.

\bibitem{stark2022-3d}
Hannes St{\"a}rk, Dominique Beaini, Gabriele Corso, Prudencio Tossou, Christian Dallago, Stephan G{\"u}nnemann, and Pietro Li{\`o}.
\newblock 3d infomax improves gnns for molecular property prediction.
\newblock {\em 39th International Conference on Machine Learning}, 2022.

\bibitem{tay2020long}
Yi~Tay, Mostafa Dehghani, Samira Abnar, Yikang Shen, Dara Bahri, Philip Pham, Jinfeng Rao, Liu Yang, Sebastian Ruder, and Donald Metzler.
\newblock Long range arena: A benchmark for efficient transformers.
\newblock In {\em International Conference on Learning Representations}, 2020.

\bibitem{tonshoff2023did}
Jan T{\"o}nshoff, Martin Ritzert, Eran Rosenbluth, and Martin Grohe.
\newblock Where did the gap go? reassessing the long-range graph benchmark.
\newblock {\em arXiv preprint arXiv:2309.00367}, 2023.

\bibitem{vaswani_2017_attention}
Ashish Vaswani, Noam Shazeer, Niki Parmar, Jakob Uszkoreit, Llion Jones, Aidan~N Gomez, {\L}ukasz Kaiser, and Illia Polosukhin.
\newblock Attention is all you need.
\newblock {\em Advances in neural information processing systems}, 30, 2017.

\bibitem{velikovic2017gat}
Petar Veli{\v{c}}kovi{\'c}, Guillem Cucurull, Arantxa Casanova, Adriana Romero, Pietro Lio, and Yoshua Bengio.
\newblock Graph attention networks.
\newblock In {\em International Conference on Learning Representations}, 2018.

\bibitem{weisfeiler1968reduction}
Boris Weisfeiler and Andrei Leman.
\newblock The reduction of a graph to canonical form and the algebra which appears therein.
\newblock {\em NTI, Series}, 2(9):12--16, 1968.

\bibitem{wu2022graph}
Lingfei Wu, Peng Cui, Jian Pei, Liang Zhao, and Le~Song.
\newblock Graph neural networks.
\newblock In {\em Graph Neural Networks: Foundations, Frontiers, and Applications}, pages 27--37. Springer, 2022.

\bibitem{xu2018powerful_gin}
Keyulu Xu, Weihua Hu, Jure Leskovec, and Stefanie Jegelka.
\newblock How powerful are graph neural networks?
\newblock In {\em International Conference on Learning Representations}, 2019.

\bibitem{LearningGenerativeDyn_MinkaiXu}
Minkai Xu, Shitong Luo, Yoshua Bengio, Jian Peng, and Jian Tang.
\newblock Learning neural generative dynamics for molecular conformation generation.
\newblock In {\em International Conference on Learning Representations}, 2021.

\bibitem{yang2021graph}
Yongyi Yang, Tang Liu, Yangkun Wang, Jinjing Zhou, Quan Gan, Zhewei Wei, Zheng Zhang, Zengfeng Huang, and David Wipf.
\newblock Graph neural networks inspired by classical iterative algorithms.
\newblock In {\em International Conference on Machine Learning}, pages 11773--11783. PMLR, 2021.

\bibitem{ying2021transformers}
Chengxuan Ying, Tianle Cai, Shengjie Luo, Shuxin Zheng, Guolin Ke, Di~He, Yanming Shen, and Tie-Yan Liu.
\newblock Do transformers really perform badly for graph representation?
\newblock {\em Advances in Neural Information Processing Systems}, 34, 2021.

\bibitem{ying2021graphormer}
Chengxuan Ying, Tianle Cai, Shengjie Luo, Shuxin Zheng, Guolin Ke, Di~He, Yanming Shen, and Tie-Yan Liu.
\newblock Do transformers really perform badly for graph representation?
\newblock In {\em Advances in Neural Information Processing Systems}, 2021.

\bibitem{you2020design}
Jiaxuan You, Rex Ying, and Jure Leskovec.
\newblock Design space for graph neural networks.
\newblock In {\em Advances in Neural Information Processing Systems}, 2020.

\bibitem{zhang2021nested}
Muhan Zhang and Pan Li.
\newblock Nested graph neural networks.
\newblock {\em Advances in Neural Information Processing Systems}, 34, 2021.

\end{thebibliography}
\bibliographystyle{plain}

\newpage
\section*{Checklist}

\begin{enumerate}

\item For all authors...
\begin{enumerate}
  \item Do the main claims made in the abstract and introduction accurately reflect the paper's contributions and scope?
    \answerYes{}
  \item Did you describe the limitations of your work?
    \answerYes{See Section~\ref{sec:characterizing}.}
  \item Did you discuss any potential negative societal impacts of your work?
    \answerNo{The proposed benchmarking suite aims at evaluating graph representation learning and does not have immediate societal impacts.}
  \item Have you read the ethics review guidelines and ensured that your paper conforms to them?
    \answerYes{}
\end{enumerate}

\item If you are including theoretical results...
\begin{enumerate}
  \item Did you state the full set of assumptions of all theoretical results?
    \answerNA{}
        \item Did you include complete proofs of all theoretical results?
    \answerNA{}
\end{enumerate}

\item If you ran experiments (e.g. for benchmarks)...
\begin{enumerate}
  \item Did you include the code, data, and instructions needed to reproduce the main experimental results (either in the supplemental material or as a URL)?
    \answerYes{Yes, the source code (with hyperlinks to automatically download the datasets) is available in the supplemental material.}
  \item Did you specify all the training details (e.g., data splits, hyperparameters, how they were chosen)?
    \answerYes{We discuss splits in the main text (Section~\ref{sec:proposed_datasets} and hyperparameters in Appendix~\ref{app:exp_details}.}
        \item Did you report error bars (e.g., with respect to the random seed after running experiments multiple times)?
    \answerYes{All experiments were executed 4 times with different random seeds, mean and standard deviation is reported.}
        \item Did you include the total amount of compute and the type of resources used (e.g., type of GPUs, internal cluster, or cloud provider)?
    \answerYes{See Appendix~\ref{app:exp_details}.}
\end{enumerate}

\item If you are using existing assets (e.g., code, data, models) or curating/releasing new assets...
\begin{enumerate}
  \item If your work uses existing assets, did you cite the creators?
    \answerYes{}
  \item Did you mention the license of the assets?
    \answerYes{}
  \item Did you include any new assets either in the supplemental material or as a URL?
    \answerYes{Available in the supplemental material and will be openly published.}
  \item Did you discuss whether and how consent was obtained from people whose data you're using/curating?
    \answerNA{}
  \item Did you discuss whether the data you are using/curating contains personally identifiable information or offensive content?
    \answerNo{Our benchmark is based on molecular and imaging datasets without personal information or offensive content.}
\end{enumerate}

\item If you used crowdsourcing or conducted research with human subjects...
\begin{enumerate}
  \item Did you include the full text of instructions given to participants and screenshots, if applicable?
    \answerNA{}
  \item Did you describe any potential participant risks, with links to Institutional Review Board (IRB) approvals, if applicable?
    \answerNA{}
  \item Did you include the estimated hourly wage paid to participants and the total amount spent on participant compensation?
    \answerNA{}
\end{enumerate}

\end{enumerate}

\clearpage
\appendix
\renewcommand\thefigure{\thesection.\arabic{figure}}
\renewcommand\thetable{\thesection.\arabic{table}}

\section{Extended Dataset Description and Results}\label{app:dataset_details}
\setcounter{figure}{0}
\setcounter{table}{0}

\subsection{Optional variants of \pascal and \coco datasets}
For \pascal and \coco datasets, the graphs that we keep by default are the \rbgraph graphs which are based on SLIC superpixels extraction with compactness value of 30. Additionally, we provide optional variants of both these datasets which are based on SLIC superpixels extraction with compactness value of 10, and two other graph construction formats, \texttt{coo} and \texttt{coo-feat}.
In this section, we include the description and results of baseline experiments of these optional datasets as well. Note that any of these version of the \texttt{SP} dataset can be used as independent LRGB dataset.

\textbf{Construction of \texttt{coo} and \texttt{coo-feat} graphs:}
Under these two construction methods, the resultant graphs are 8 nearest neighbor graphs where the pairwise adjacency weights for two nodes are \textit{first} constructed based on coordinates (for \texttt{coo}) or based on coordinates and feature intensities (for \texttt{coo-feat}) of the superpixels nodes, and \textit{then} each node is directly connected to 8 other nodes with the highest weight scores. The weights computation is based on the Eqn. \ref{eqn:coo} for \texttt{coo} and Eqn. \ref{eqn:coo-feat} for \texttt{coo-feat}:
%
  \begin{equation}
    W^{8-nn}_{ij} = \textrm{exp}\left(- \frac{\|x_i - x_j \|}{\sigma^2_x} \right) \ \label{eqn:coo}
  \end{equation}
  \begin{equation}
    W^{8-nn}_{ij} = \textrm{exp}\left(- \frac{\|x_i - x_j \|}{\sigma^2_x} - \frac{\|f_i - f_j \|}{\sigma^2_f} \right) \  \label{eqn:coo-feat}
  \end{equation}

    where, $x_i, x_j$ are the 2 dimensional coordinates, and $f_i, f_j$ are the 12 dimensional (3 dimensions each of mean, std, max, min) RGB feature values of superpixels $i, j$ respectively, $\sigma^2_x$ is a scaling parameter defined as the average distance $x_k$ of the $k=8$ nearest neighbors for each node.
The initial feature of each superpixel node is 12 dimensional RGB feature value (mean, std, max, min) and that of an edge between two nodes is a 1 dimensional edge weight that is given by Eqn. \ref{eqn:coo} or \ref{eqn:coo-feat} for the respective graph format.

\textbf{Statistics and Baseline Results.} The complete statistics of the aforementioned optional versions of the \pascal and \coco datasets are included in Table \ref{tab:data_summary_extended}. Note that that versions with the options \textbf{SLIC}: 30 and \textbf{Graph}: \rbgraph is the default dataset for both \pascal and \coco that we present in the main Section \ref{sec:proposed_datasets}. The results of the baseline experiments on all the dataset versions are reported in Table \ref{tab:experiments_superpixels_full}.

\begin{table}[ht]
  \caption{Statistics of 6 tried versions of \pascal and \coco datasets, each derived with a different combination of \textbf{Options} in terms of: (i) \textbf{SLIC}, which denotes the value of \texttt{compactness} parameter used during the extraction of superpixels by SLIC algorithm \cite{achanta2010slic} and (ii) \textbf{Graph}, which denotes the graph format that was used to construct the adjacency matrix. The \textbf{Graph} options `\texttt{coo}' refers to a 8-nn graph where the edge weight is based on superpixel coordinates (Eqn. \ref{eqn:coo}),  `\texttt{coo-feat}' refers to a 8-nn graph where the edge weight is based on superpixel coordinates as well as feature intensities (Eqn. \ref{eqn:coo-feat}), `\rbgraph' refers to a region boundary graph.
    }
    \begin{adjustwidth}{-2.5 cm}{-2.5 cm}\centering
    \scalebox{0.7}{
    \begin{tabular}{l c c r ccc cc cc}
    \toprule
    \multirow{2}{*}{\textbf{Dataset}} & 
    \multicolumn{2}{c}{\textbf{Options}} & \multirow{2}{*}{\shortstack[c]{\textbf{Total} \\ \textbf{Graphs}}} & \multirow{2}{*}{\shortstack[c]{\textbf{Total} \\ \textbf{Nodes}}} & \multirow{2}{*}{\shortstack[c]{\textbf{Avg} \\ \textbf{Nodes}}} & 
    \multirow{2}{*}{\shortstack[c]{\textbf{Mean} \\\textbf{Deg.}}} &
    \multirow{2}{*}{\shortstack[c]{\textbf{Total} \\ \textbf{Edges}}} &
    \multirow{2}{*}{\shortstack[c]{\textbf{Avg} \\ \textbf{Edges}}} & 
    \multirow{2}{*}{\shortstack[c]{\textbf{Avg} \\ \textbf{Short.Path.}}} &
    \multirow{2}{*}{\shortstack[c]{\textbf{Avg} \\ \textbf{Diameter}}}\\
    & \textbf{SLIC} & \textbf{Graph} & & & & & & & &\\ 
      \midrule
    \multirow{6}{*}{\pascal} & 
    \multirow{3}{*}{10} & \texttt{coo} & \multirow{3}{*}{11,355} & \multirow{3}{*}{4,747,374} & \multirow{3}{*}{418.09} & 8.00 & 37,978,992 & 3,344.69 & 7.50$\pm$0.76 & 17.89$\pm$2.10 \\
    & & \texttt{coo-feat} &  &  &  & 8.00 & 37,978,992 & 3,344.69 & 7.50$\pm$0.76 & 17.89$\pm$2.10\\
    & & \texttt{reg-bound} &  &  &  & 5.62 & 26,659,158 & 2,347.79 & 9.08$\pm$1.23 & 22.99$\pm$3.70 \\
    \cmidrule{2-11}
    & \multirow{3}{*}{30} & \texttt{coo} & \multirow{3}{*}{11,355} & \multirow{3}{*}{5,443,545} & \multirow{3}{*}{479.40} & 8.00 & 43,548,360 & 3,835.17 & 8.05$\pm$0.18 & 19.40$\pm$0.65 \\
    & & \texttt{coo-feat} & & & & 8.00 & 43,548,360 & 3,835.17 & 8.05$\pm$0.18 & 19.40$\pm$0.65\\
    & & \texttt{reg-bound} & & & & 5.65 & 30,777,444 & 2,710.48 & 10.74$\pm$0.51 & 27.62$\pm$2.13\\
    \midrule
    \multirow{6}{*}{\coco} & 
    \multirow{3}{*}{10} & \texttt{coo} & \multirow{3}{*}{123,286} & \multirow{3}{*}{49,732,322} & \multirow{3}{*}{403.39} & 8.00 & 397,858,524 & 3,227.12 & 7.39$\pm$0.77 & 17.61$\pm$2.12\\
    & & \texttt{coo-feat} & & & & 8.00 & 397,858,524 & 3,227.12 & 7.39$\pm$0.77 & 17.61$\pm$2.12 \\
    & & \texttt{reg-bound} & & & & 5.61 & 278,816,918 & 2,261.55 & 8.85$\pm$1.23 & 22.40$\pm$3.70\\
    \cmidrule{2-11}
    & \multirow{3}{*}{30} & \texttt{coo} & \multirow{3}{*}{123,286} & \multirow{3}{*}{58,793,216} & \multirow{3}{*}{476.88} & 8.00 & 470,345,728 & 3,815.08 & 8.06$\pm$0.18 & 19.42$\pm$0.70\\
    & & \texttt{coo-feat} & & & & 8.00 & 470,345,728 & 3,815.08 & 8.06$\pm$0.18 & 19.42$\pm$0.70\\
    & & \texttt{reg-bound} & & & & 5.65 & 332,091,902 & 2,693.67 & 10.66$\pm$0.55 & 27.39$\pm$2.14\\
    \bottomrule
    \end{tabular}
    }
    \label{tab:data_summary_extended}
  \end{adjustwidth}
\end{table} 
\clearpage

\begin{table}[h]
    \caption{Baseline experiments for \pascal and \coco for node classification task. Performance metric is macro F1 on the respective splits (Higher is better). All experiments are run 4 times with 4 different seeds. All models have approximately 500k learnable parameters for fair comparison. The MP-GNN models are 8 layers deep, while the transformer-based models have 4 layers in order to maintain comparable hidden representation size at the fixed parameter budget.
    }
    \label{tab:experiments_superpixels_full}
    \begin{adjustwidth}{-2.5 cm}{-2.5 cm}\centering
    \scalebox{0.72}{
    \setlength\tabcolsep{3pt} 
    \begin{tabular}{c c l c A B A B A B}
    \toprule
    \multicolumn{2}{c}{\multirow{2}{*}{\textbf{\thead{Dataset \\ (SLIC)}}}} & \multirow{2}{*}{\textbf{Model}} & \multirow{2}{*}{\hspace*{-2em}\textbf{\# Params}} & \multicolumn{2}{c}{\texttt{coo}} & \multicolumn{2}{c}{\texttt{coo-feat}} & \multicolumn{2}{c}{\texttt{reg-bound}}\\ \cmidrule(lr){5-6} \cmidrule(lr){7-8} \cmidrule(lr){9-10}
    & & & & \textbf{Train F1} & \textbf{Test F1 $\uparrow$} & \textbf{Train F1} & \textbf{Test F1 $\uparrow$} & \textbf{Train F1} & \textbf{Test F1 $\uparrow$}\\
    \midrule
    \parbox[t]{5mm}{\multirow{12}{*}{\rotatebox[origin=c]{90}{\pascal}}}& \parbox[t]{2mm}{\multirow{6}{*}{\rotatebox[origin=c]{90}{\textbf{10}}}}& GCN &496k &0.1559$\pm$0.0079 &0.1281$\pm$0.0025 &0.1956$\pm$0.0202 &0.1321$\pm$0.0043 &0.1530$\pm$0.0048 &0.1306$\pm$0.0025 \\
    & & GINE &505k &0.2178$\pm$0.0382 &0.1127$\pm$0.0039 &0.3007$\pm$0.0461 &0.1078$\pm$0.0035 &0.2278$\pm$0.0224 &0.1231$\pm$0.0052 \\
    & & GatedGCN & 502k & 0.4319$\pm$0.0187 & 0.2788$\pm$0.0032 & 0.3560$\pm$0.0567 & 0.2289$\pm$0.0137 & 0.3574$\pm$0.0573 & 0.2705$\pm$0.0251\\
    & & GatedGCN+LapPE &502k &0.4390$\pm$0.0144 &0.2803$\pm$0.0031 &0.3535$\pm$0.0376 &0.2241$\pm$0.0035 &0.3553$\pm$0.0396 &0.2722$\pm$0.0149 \\
    & & Transformer+LapPE &501k &0.6140$\pm$0.0635 &0.2661$\pm$0.0129 &0.5594$\pm$0.0445 &0.2667$\pm$0.0060 &0.5925$\pm$0.0447 &0.2627$\pm$0.0086 \\
    & & SAN+LapPE &531k &0.6691$\pm$0.0339 &0.2904$\pm$0.0031 &0.5555$\pm$0.0650 &0.2808$\pm$0.0047 &0.5636$\pm$0.0506 &0.3031$\pm$0.0046 \\
    & & SAN+RWSE &468k &0.6200$\pm$0.0502 &0.2841$\pm$0.0090 &0.5726$\pm$0.0615 &0.2764$\pm$0.0184 &0.5968$\pm$0.0487 &0.3113$\pm$0.0072 \\
    \cmidrule{2-10}
    & \parbox[t]{2mm}{\multirow{6}{*}{\rotatebox[origin=c]{90}{\textbf{30}}}}& GCN &496k &0.1469$\pm$0.0068 &0.1262$\pm$0.0031 &0.1742$\pm$0.0042 &0.1326$\pm$0.0015 &0.1450$\pm$0.0125 &0.1268$\pm$0.0060 \\
    & & GINE &505k &0.2575$\pm$0.0283 &0.1203$\pm$0.0045 &0.2479$\pm$0.0318 &0.1035$\pm$0.0015 &0.2088$\pm$0.0268 &0.1265$\pm$0.0076 \\
    & & GatedGCN & 502k & 0.4311$\pm$0.0325 & 0.2916$\pm$0.0058 & 0.3379$\pm$0.0107 & 0.2410$\pm$0.0015 & 0.3552$\pm$0.0451 &  0.2873$\pm$0.0219\\
    & & GatedGCN+LapPE &502k &0.4223$\pm$0.0356 &0.2890$\pm$0.0057 &0.3110$\pm$0.0706 &0.2317$\pm$0.0217 &0.3512$\pm$0.0167 &0.2860$\pm$0.0085 \\
    & & Transformer+LapPE &501k &0.6213$\pm$0.0393 &0.2633$\pm$0.0056 &0.6607$\pm$0.0684 &0.2697$\pm$0.0081 &0.7170$\pm$0.0048 &0.2694$\pm$0.0098 \\
    & & SAN+LapPE &531k &0.6485$\pm$0.0711 &0.3218$\pm$0.0160 &0.5242$\pm$0.0480 &0.3003$\pm$0.0046 &0.5723$\pm$0.0427 &0.3230$\pm$0.0039 \\
    & & SAN+RWSE &468k &0.6240$\pm$0.0866 &0.3227$\pm$0.0084 &0.5869$\pm$0.0349 &0.3124$\pm$0.0091 &0.5819$\pm$0.0331 &0.3216$\pm$0.0027 \\
    \midrule[1pt]
    \parbox[t]{5mm}{\multirow{12}{*}{\rotatebox[origin=c]{90}{\coco}}}& \parbox[t]{2mm}{\multirow{6}{*}{\rotatebox[origin=c]{90}{\textbf{10}}}}& GCN &509k &0.0852$\pm$0.0030 &0.0770$\pm$0.0017 &0.0919$\pm$0.0058 &0.0780$\pm$0.0026 &0.0885$\pm$0.0078 &0.0809$\pm$0.0043 \\
    & & GINE &515k &0.1874$\pm$0.0071 &0.1109$\pm$0.0048 &0.1605$\pm$0.0090 &0.0846$\pm$0.0045 &0.1812$\pm$0.0155 &0.1196$\pm$0.0053 \\
    & & GatedGCN &508k &0.3009$\pm$0.0078 &0.2280$\pm$0.0032 &0.2842$\pm$0.0077 &0.2130$\pm$0.0036 &0.3149$\pm$0.0099 &0.2542$\pm$0.0044 \\
    & & GatedGCN+LapPE &509k &0.3018$\pm$0.0057 &0.2307$\pm$0.0014 &0.2789$\pm$0.0080 &0.2110$\pm$0.0036 &0.3184$\pm$0.0144 &0.2529$\pm$0.0063 \\
    & & Transformer+LapPE &508k &0.3700$\pm$0.0141 &0.2455$\pm$0.0036 &0.3775$\pm$0.0082 &0.2492$\pm$0.0036 &0.3758$\pm$0.0205 &0.2478$\pm$0.0068 \\
    & & SAN+LapPE &536k &0.3437$\pm$0.0096 &0.2605$\pm$0.0062 &0.3278$\pm$0.0041 &0.2596$\pm$0.0015 &0.2541$\pm$0.0394 &0.2325$\pm$0.0191 \\
    & & SAN+RWSE &474k &0.3557$\pm$0.0264 &0.2675$\pm$0.0126 &0.3270$\pm$0.0145 &0.2585$\pm$0.0046 &0.2815$\pm$0.0371 &0.2442$\pm$0.0231 \\
    \cmidrule{2-10}
    & \parbox[t]{2mm}{\multirow{6}{*}{\rotatebox[origin=c]{90}{\textbf{30}}}}& GCN &509k &0.0914$\pm$0.0056 &0.0797$\pm$0.0026 &0.1003$\pm$0.0043 &0.0843$\pm$0.0019 &0.0948$\pm$0.0014 &0.0841$\pm$0.0010 \\
    & & GINE &515k &0.1742$\pm$0.0186 &0.1168$\pm$0.0053 &0.1646$\pm$0.0081 &0.1003$\pm$0.0022 &0.2100$\pm$0.0041 &0.1339$\pm$0.0044 \\
    & & GatedGCN &508k &0.3024$\pm$0.0043 &0.2441$\pm$0.0035 &0.2926$\pm$0.0154 &0.2285$\pm$0.0069 &0.3167$\pm$0.0059 &0.2641$\pm$0.0045 \\
    & & GatedGCN+LapPE &509k &0.3101$\pm$0.0062 &0.2454$\pm$0.0015 &0.2894$\pm$0.0060 &0.2283$\pm$0.0036 &0.3102$\pm$0.0112 &0.2574$\pm$0.0034 \\
    & & Transformer+LapPE &508k &0.3855$\pm$0.0185 &0.2579$\pm$0.0057 &0.3750$\pm$0.0224 &0.2589$\pm$0.0069 &0.3912$\pm$0.0098 &0.2618$\pm$0.0031 \\
    & & SAN+LapPE &536k &0.3376$\pm$0.0455 &0.2781$\pm$0.0143 &0.2941$\pm$0.0810 &0.2498$\pm$0.0513 &0.2830$\pm$0.0246 &0.2592$\pm$0.0158 \\
    & & SAN+RWSE &474k &0.3652$\pm$0.0104 &0.2817$\pm$0.0047 &0.3754$\pm$0.0074 &0.2869$\pm$0.0067 &0.2657$\pm$0.0224 &0.2434$\pm$0.0156 \\
    \bottomrule
    \end{tabular}
    }
    \end{adjustwidth}
\end{table}

\subsection{Extended description of \pepstruct}
\label{app:peptstruct_description}

Below, we describe the 11 tasks from the \pepstruct dataset, which represented properties computed from the 3D structure, then normalized to zero mean and unit standard deviation.
\vspace{-3pt}
\begin{itemize}[leftmargin=1.5em]
    \vspace{-3pt}
    \item \textbf{Inertia\_mass} The inertia of the molecules according to its 3 principal components, using the mass of the \underline{atoms} and their distances to the centroid.
    \vspace{-3pt}
    \item \textbf{Inertia\_valence} The inertia of the molecules according to its 3 principal components, using the \underline{valence} of the atoms and their distances to the centroid.
    \vspace{-3pt}
    \item \textbf{Length} The maximum distance between each atom-pairs in each of its 3 main axes.
    \vspace{-3pt}
    \item \textbf{Sphericity} A measure of how much the molecule looks like a sphere: the ratio of the molecule's surface area to the surface area of a sphere with similar volume.
    \vspace{-3pt}
    \item \textbf{Plane\_best\_fit} The average distance of all heavy atoms from the plane of best fit.
\end{itemize}

\subsection{Extended Results for \pepstruct}
\label{app:extended_results}

\begin{table}[ht]
    \caption{Extended evaluation metrics for \pepstruct. The training and testing performance is quantified in terms of the Coefficient of Determination $\mathrm{R}^2$, in addition to MAE reported in Table~\ref{tab:experiments_peptides}.
    }
    \label{tab:pep_r2}
    \centering
    \scalebox{0.85}{
    \setlength\tabcolsep{4pt} 
    \begin{tabular}{l r A C A B}\toprule
    \textbf{Model} &\hspace*{-1em}\textbf{\# Params.} &\textbf{Train MAE} &\textbf{Test MAE $\downarrow$} &\textbf{Train R2} &\textbf{Test R2 $\uparrow$} \\\midrule
    GCN &508k &0.2939$\pm$0.0055 &0.3496$\pm$0.0013 &0.6513$\pm$0.0078 &0.6019$\pm$0.0027 \\
    GCNII & 505k & 0.2957$\pm$0.0025 & 0.3471$\pm$0.0010 & 0.6913$\pm$0.0242 & 0.6148$\pm$0.0054\\
    GINE &476k &0.3116$\pm$0.0047 &0.3547$\pm$0.0045 &0.6494$\pm$0.0108 &0.5943$\pm$0.0067 \\
    GatedGCN &509k &0.2761$\pm$0.0032 &0.3420$\pm$0.0013 &0.6907$\pm$0.0058 &0.6254$\pm$0.0013 \\
    GatedGCN+RWSE &506k &0.2578$\pm$0.0116 &0.3357$\pm$0.0006 &0.7204$\pm$0.0149 &0.6329$\pm$0.0034 \\\midrule
    Transformer+LapPE &488k &0.2403$\pm$0.0066 &\first{0.2529$\pm$0.0016} &0.8027$\pm$0.0108 &\first{0.7743$\pm$0.0053} \\
    SAN+LapPE &493k &0.2822$\pm$0.0108 &0.2683$\pm$0.0043 &0.6887$\pm$0.0153 &0.7581$\pm$0.0057 \\
    SAN+RWSE &500k &0.2680$\pm$0.0038 &0.2545$\pm$0.0012 &0.7112$\pm$0.0049 &0.7716$\pm$0.0034 \\
    \bottomrule
    \end{tabular}
    }
\end{table}

\subsection{Correlation of labels in \pepfunc and \pepstruct}

\begin{figure}[ht]
    \centering
    \includegraphics[trim=20 5 20 2, width=1\textwidth]{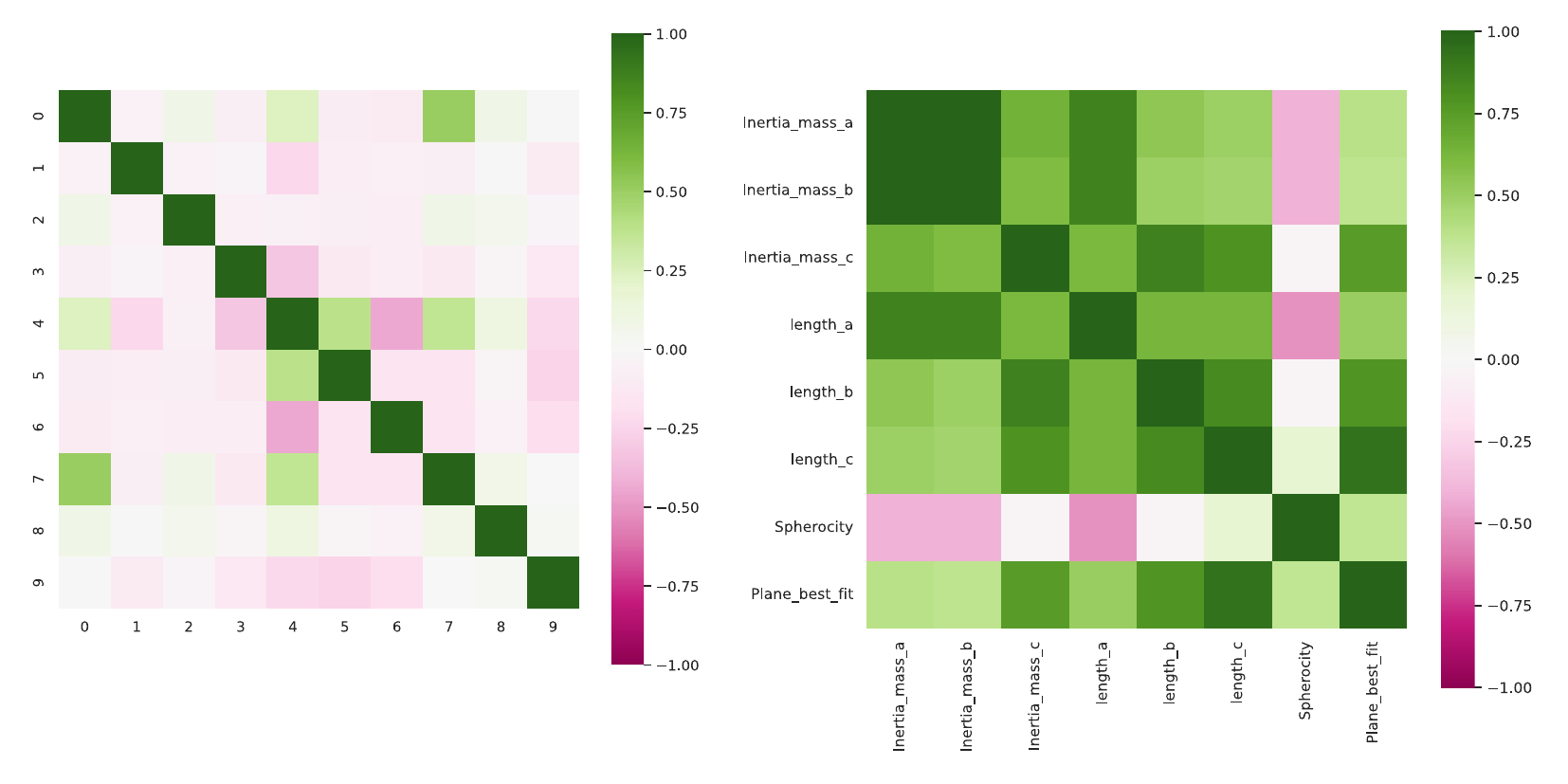}
\caption{\textbf{Left:} Visualization of pearson correlation between classes in peptides-func dataset. \textbf{Right:} Visualization of correlation between geometric properties in peptides-struct dataset.}
\label{fig:pep_clas_corr}
\end{figure}

The \pepfunc is a multi-label classification and \pepstruct is a multi-label regression. Evaluating the correlation between the labels will ensure that labels are not redundant and provide a variety of information, Figure \ref{fig:pep_clas_corr}. We observe that there is very little correlation between classes of peptides. For the structural dataset, there are some correlations, which are expected since intertia is related to length and spherocity, but in general, the correlation remains limited, which motivates using a multi-label regression.

\subsection{Dataset Licenses.}
The information on the dataset sources that we used for the proposed LRGB datasets' preparation, the original licenses of use and the release licenses are in Table \ref{tab:licenses}.

\begin{table}[ht]
\caption{Original resources that our 5 datasets are derived from and their licensing information. *Custom License for Pascal VOC 2011 (respecting Flickr terms of use).}\label{tab:licenses}
\centering
\small
\begin{tabular}{lrrrr}\toprule
&Derived from &Original License &Release License \\\midrule
\pascal &Pascal VOC \cite{everingham2010pascal} & Custom* & Custom* \\
\coco &MS COCO \cite{lin2014microsoft} &CC BY 4.0 &CC BY 4.0 \\
\pcqmcontact &PCQM4Mv2 \cite{hu2021ogb_lsc} &CC BY 4.0 &CC BY 4.0 \\
\pepfunc &SATPdb \cite{singh2016satpdb} &CC BY-NC 4.0 &CC BY-NC 4.0 \\
\pepstruct &SATPdb \cite{singh2016satpdb} &CC BY-NC 4.0 &CC BY-NC 4.0 \\
\bottomrule
\end{tabular}
\end{table}

\newpage
\section{Visualizations} \label{sec:viz_final_sp}
\setcounter{figure}{0}
\setcounter{table}{0}

\subsection{\pascal}
\begin{figure}[!h]
\centering
\begin{subfigure}
  \centering
  \includegraphics[width=0.99\linewidth]{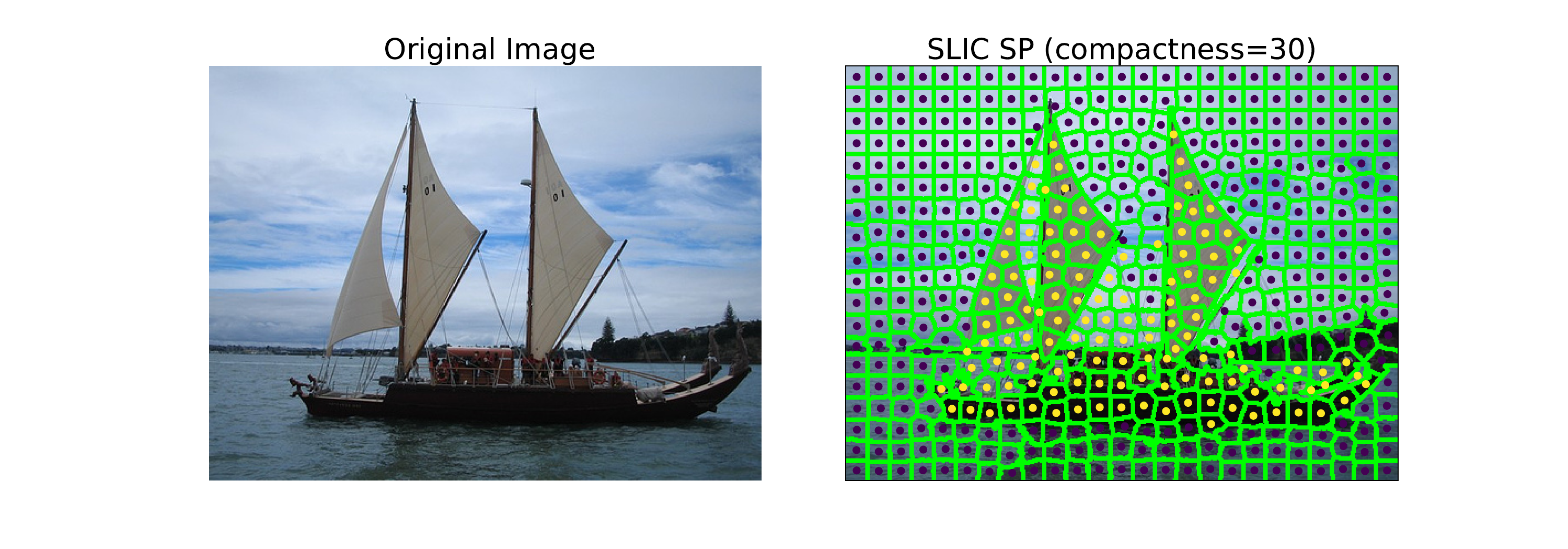}  
  \label{fig:voc-row1}
\end{subfigure}
\vspace{-20pt}
\begin{subfigure}
  \centering
  \includegraphics[width=0.99\linewidth]{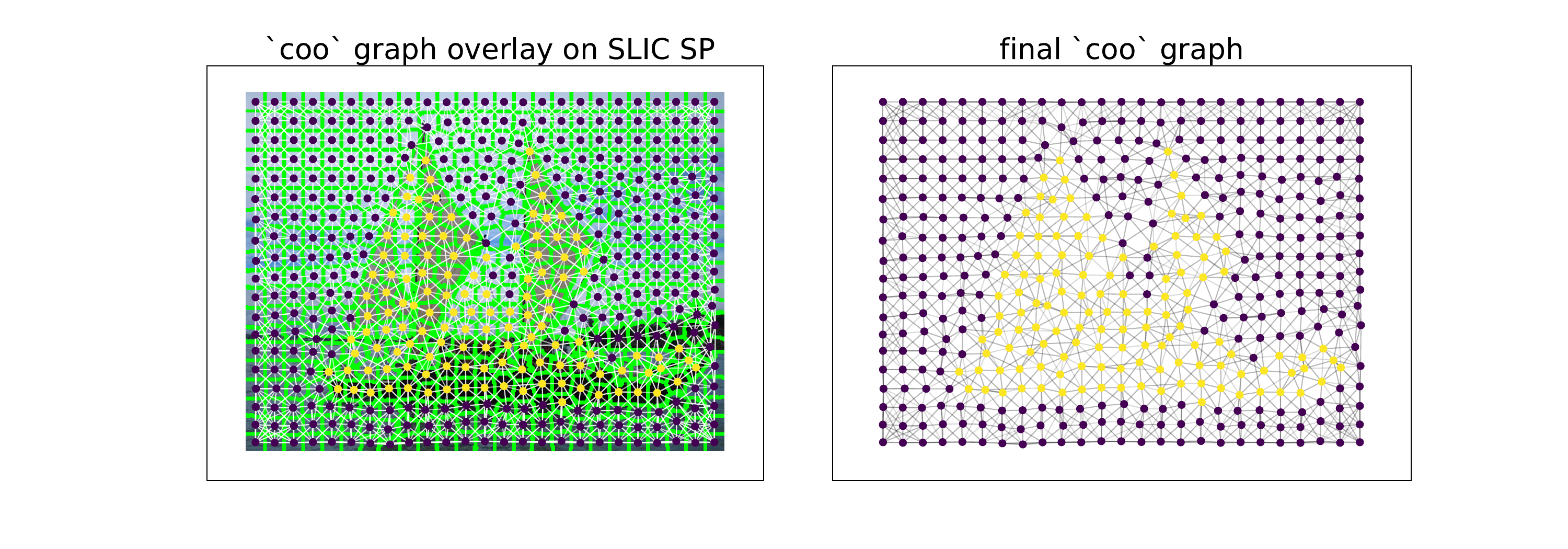}  
  \label{fig:voc-row2}
\end{subfigure}
\vspace{-20pt}
\begin{subfigure}
  \centering
  \includegraphics[width=0.99\linewidth]{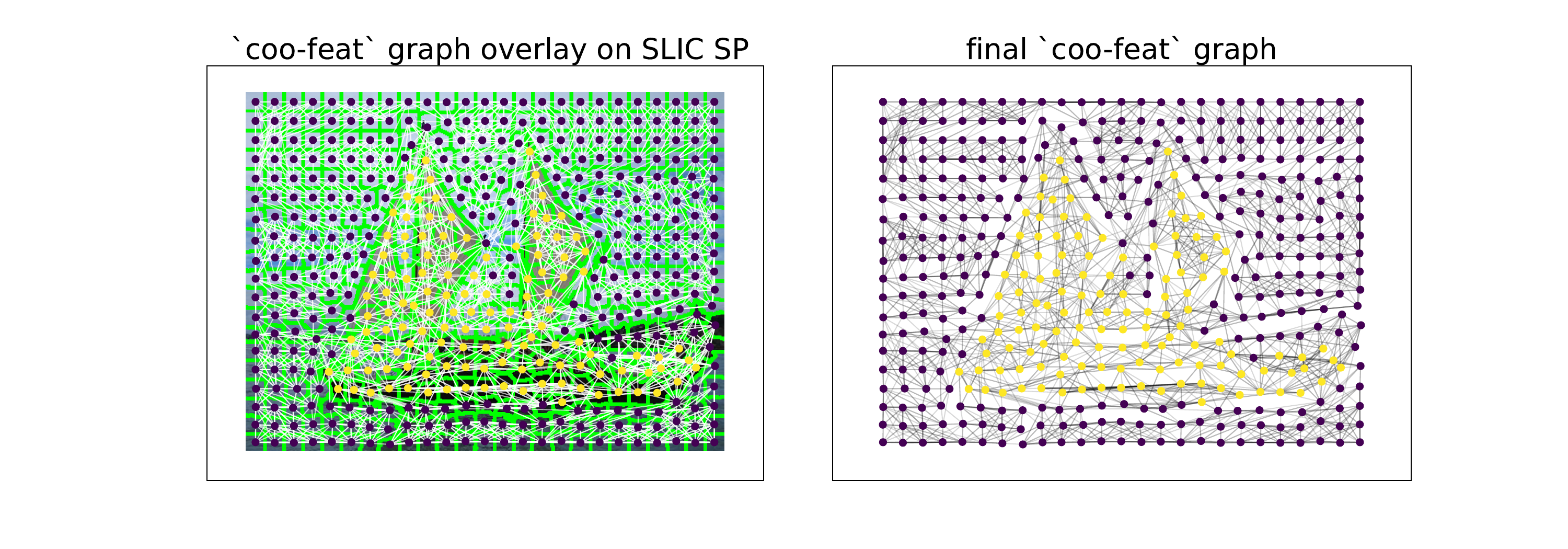}  
  \label{fig:voc-row3}
\end{subfigure}
\vspace{-20pt}
\begin{subfigure}
  \centering
  \includegraphics[width=0.99\linewidth]{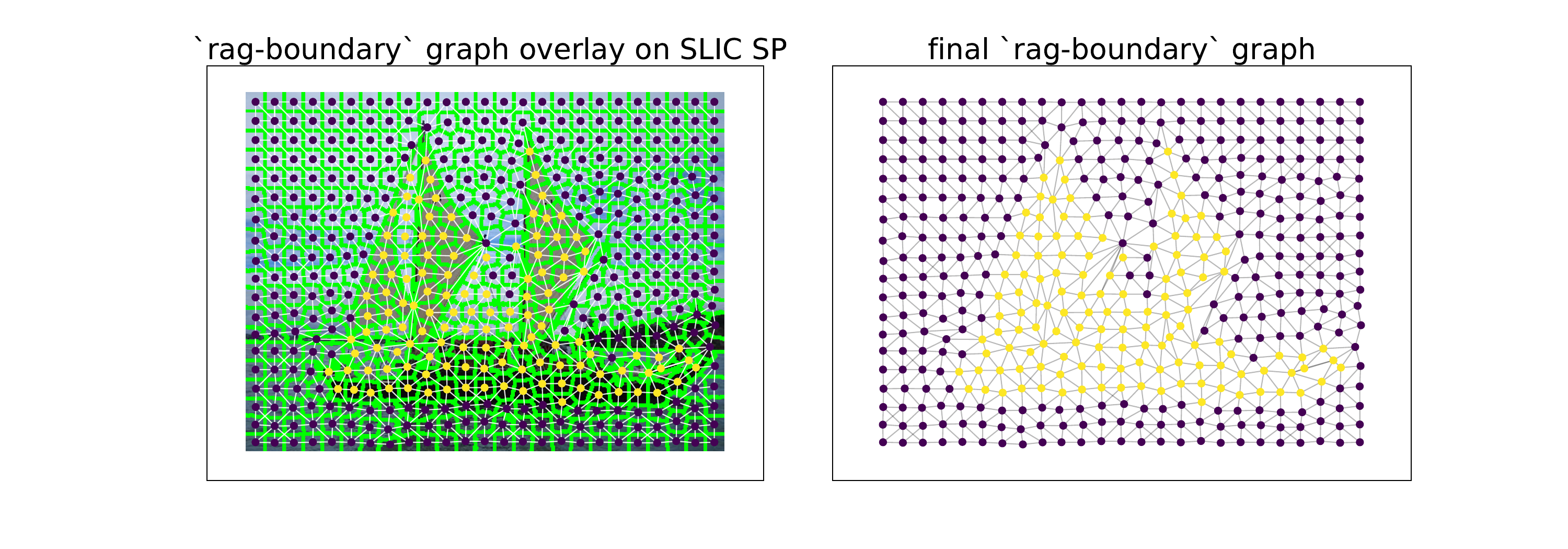}  
  \label{fig:voc-row4}
\end{subfigure}
\caption{Visualizations of a sample image and its SP graphs from \pascal dataset with 465 nodes and 3,720 edges each for \texttt{coo}, \texttt{coo-feat} graph and 2,628 edges for \texttt{reg-bound} graph. Unique colors on the nodes denote the corresponding node labels.}
\label{fig:fig_viz_voc}
\end{figure}

\newpage
\subsection{\coco}
\begin{figure}[!h]
\centering
\begin{subfigure}
  \centering
  \includegraphics[width=0.99\linewidth]{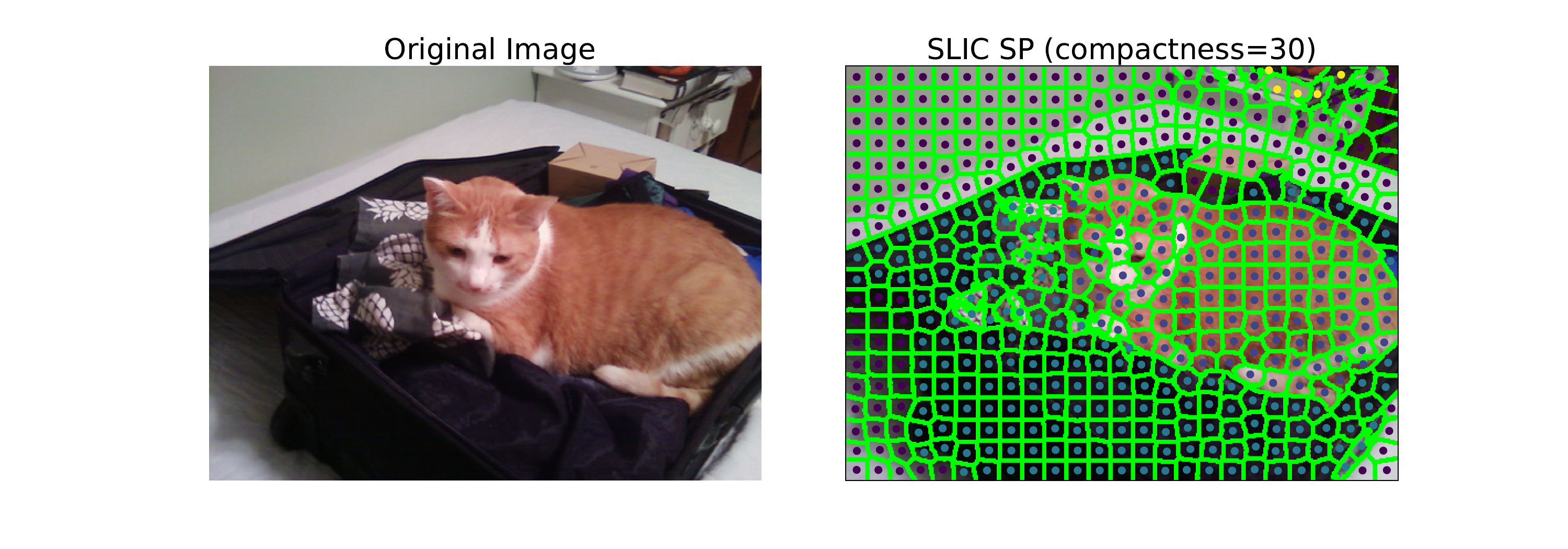}  
  \label{fig:coco-row1}
\end{subfigure}
\vspace{-20pt}
\begin{subfigure}
  \centering
  \includegraphics[width=0.99\linewidth]{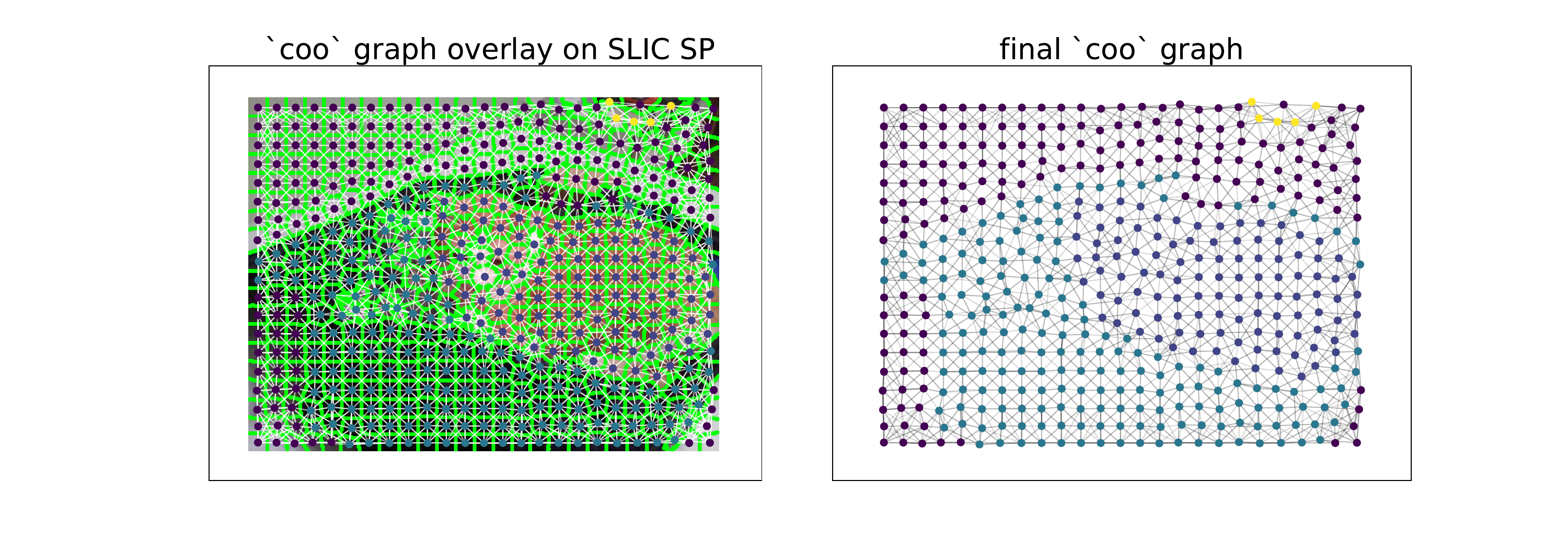}  
  \label{fig:coco-row2}
\end{subfigure}
\vspace{-20pt}
\begin{subfigure}
  \centering
  \includegraphics[width=0.99\linewidth]{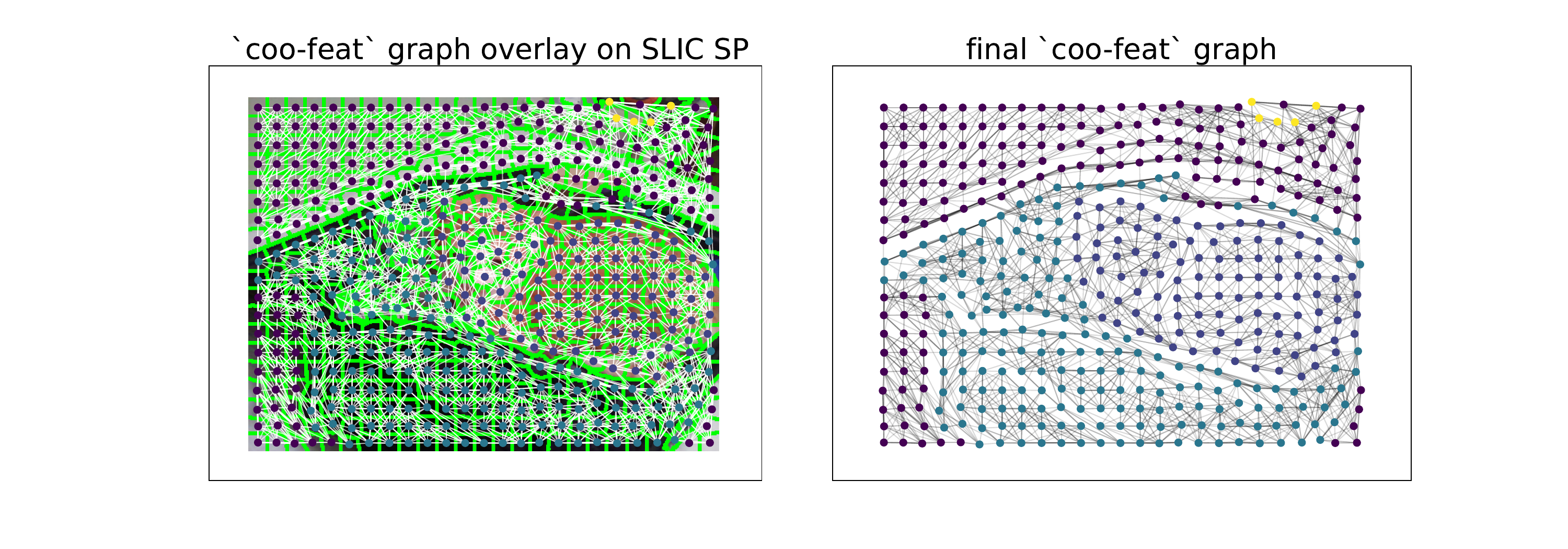}  
  \label{fig:coco-row3}
\end{subfigure}
\vspace{-20pt}
\begin{subfigure}
  \centering
  \includegraphics[width=0.99\linewidth]{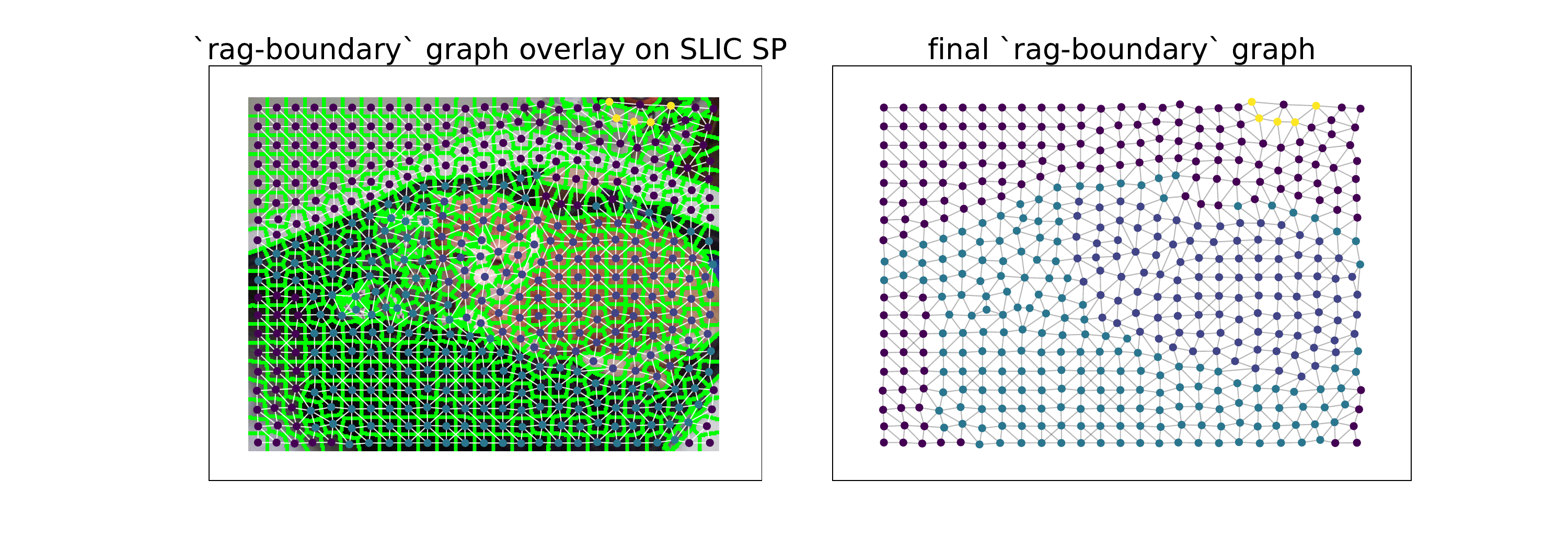}  
  \label{fig:coco-row4}
\end{subfigure}
\caption{Visualizations of a sample image and its SP graphs from \coco dataset with 470 nodes and 3,760 edges each for \texttt{coo}, \texttt{coo-feat} graph and 2,662 edges for \texttt{reg-bound} graph. Unique colors on the nodes denote the corresponding node labels.}
\label{fig:fig_viz_coco}
\end{figure}

\newpage
\subsection{\pepfunc and \pepstruct}

\begin{figure}[h]
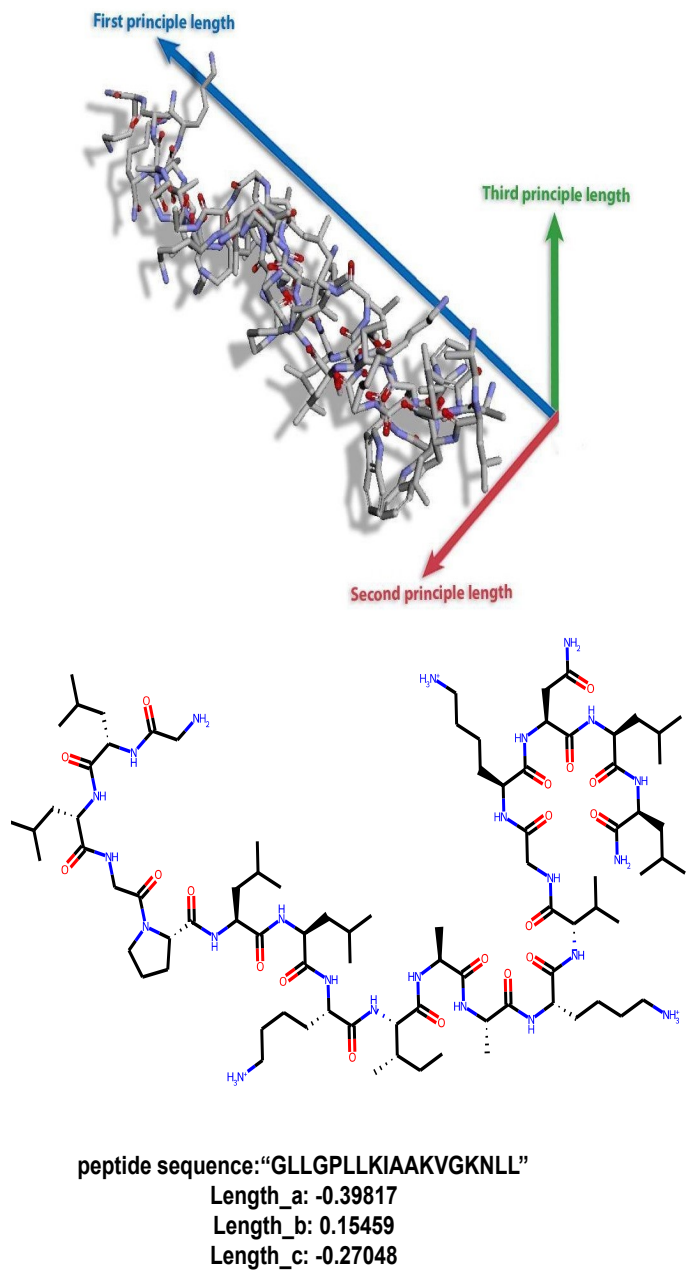

    
    \centering
    \includegraphics[trim=30 5 900 30, clip, width=0.55\textwidth]{images/peptide-3d-1.pdf}
    \vspace{40pt}
    \includegraphics[trim=840 5 30 50, clip, width=0.65\textwidth]{images/peptide-3d-1.pdf}
    \vspace{-10pt}
    \caption{Large size visualization of Figure \ref{fig:pep_3d_plot}. \textbf{Top:} 3D Visualization of "GLLGPLLKIAAKVGKNLL" peptide. \textbf{Bottom:} The molecular graph for the same peptide. }
    \label{fig:pep_3d_plot_large}
    
\end{figure}


\newpage
\section{Experimental Details} \label{app:exp_details}
\setcounter{figure}{0}
\setcounter{table}{0}

\begin{table}[ht]
    \caption{Baseline hyperparameters of 7 evaluated models on the 5 new LRGB benchmarks. Shown is the size of the hidden node representation $d$ and the number of layers $L$. Where applicable, the type of positional/structural embedding is shown: LapPE-$k$ denotes Laplacian positional encoding \cite{kreuzer2021rethinking} with first $k$ non-trivial eigenvectors (with original Transformer-based encoder for SAN~\cite{kreuzer2021rethinking} and more parameter-efficient DeepSet encoder for GatedGCN); RWSE-$m$ denotes random-walk structural encoding \cite{dwivedi2022graph} with $1..m$ steps and a linear encoder.}
    \label{tab:hparams}
    \scalebox{0.85}{
    \centering
    \setlength\tabcolsep{4pt} 
    \rowcolors{2}{lightgray}{}
    \renewcommand{\arraystretch}{1.2}
    \begin{tabular}{l c c c c c}\toprule
    &\pascal &\coco &\pcqmcontact &\pepfunc & \pepstruct \\\midrule
    GCN &$d$=220, $L$=8 &$d$=220, $L$=8 &$d$=275, $L$=5 &$d$=300, $L$=5 &$d$=300, $L$=5 \\
    GCNII &$d$=220, $L$=8 &$d$=220, $L$=8 &$d$=275, $L$=5 &$d$=300, $L$=5 &$d$=300, $L$=5 \\
    GINE &$d$=166, $L$=8 &$d$=166, $L$=8 &$d$=208, $L$=5 &$d$=208, $L$=5 &$d$=208, $L$=5 \\
    GatedGCN(+PE/SE) &$d$=108, $L$=8 &$d$=108, $L$=8 &$d$=138, $L$=5 &$d$=138, $L$=5 &$d$=138, $L$=5 \\
    \hspace{3em} used PE/SE &LapPE-10 &LapPE-10 &RWSE-16 &RWSE-16 &RWSE-16 \\
    Transformer+LapPE &$d$=120, $L$=4 &$d$=120, $L$=4 &$d$=120, $L$=4 &$d$=120, $L$=4 &$d$=120, $L$=4 \\
    &LapPE-10 &LapPE-10 &LapPE-10 &LapPE-10 &LapPE-10 \\
    SAN+LapPE &$d$=88, $L$=4 &$d$=88, $L$=4 &$d$=84, $L$=4 &$d$=84, $L$=4 &$d$=84, $L$=4 \\
    &LapPE-10 &LapPE-10 &LapPE-10 &LapPE-10 &LapPE-10 \\
    SAN+RWSE &$d$=96, $L$=4 &$d$=96, $L$=4 &$d$=100, $L$=4 &$d$=100, $L$=4 &$d$=100, $L$=4 \\
    &RWSE-16 &RWSE-16 &RWSE-16 &RWSE-16 &RWSE-16 \\
    \bottomrule
    \end{tabular}
    }
\end{table}

\subsection{Details on Baseline Experiments Setup}

\textbf{Models.} We use GCN \cite{kipf2016gcn}, GCNII~\cite{chen2020GCNII}, GINE \cite{xu2018powerful_gin, hu2019strategies} and GatedGCN \cite{bresson2017gatedGCN} models from the local MP-GNNs class, and fully connected Transformer \cite{vaswani_2017_attention} with Laplacian PE (LapPE) \cite{dwivedi2020benchmarking, dwivedi2020generalization} and SAN \cite{kreuzer2021rethinking} models among the Transformer class. The GCN (Graph Convolutional Network) is the most popularly used local MP-GNN baseline, GCNII~\cite{chen2020GCNII} is an extension of the vanilla GCN, GINE (Graph Isomorphism Network) is a 1-WL expressive MP-GNN with ability to incorporate edge features into its update equation \cite{hu2019strategies}, and GatedGCN (Gated Graph Convolutional Network) is a soft-attention based GCN which uses learned edge gates to improve the aggregation procedure. Transformer with LapPE is a generalization of the vanilla Transformer network \cite{vaswani_2017_attention} from Natural Language Processing (NLP) domain to graphs and SAN (Spectral Attention Network) is a powerful fully-connected Graph Transformer which includes a learned PE module based on Laplacian eigenvectors and eigenvalues, alongside separate treatment of real and non-real graph edges \cite{kreuzer2021rethinking}. We use SAN with LapPE as well RWSE (Random Walk Structural Encoding) \cite{dwivedi2022graph}. The collection of above baseline models allows us to show performance trends using simple, straightforward models such as GCN and Transformer to advanced ones such as GatedGCN and SAN. We believe this  baseline collection, albeit small, represents a diverse representation of the course of action graph deep learning has evolved to, reaching at a stage where we can embark conveniently towards the development of GNNs that learn efficiently to propagate long-range dependencies. 

\textbf{Experimental Setup.} In order to facilitate fair comparison and reliable discussion of the observed trends, we select the hyperparameters of the aforementioned baselines such that they yield models within a budget of approx. 500k learnable parameters. To this end, we configure 4-8 layers deep models and adjust their hidden dimension size accordingly to the 500k parameter budget. For a list of hyperparameters used in each baseline see Table \ref{tab:hparams}. We run each experiment 4 times with different random seeds and report the mean and standard deviation of the respective performance metrics.

For optimization, we use Adam~\cite{kingma2017adam} with default settings. We set the starting learning rate between 0.0003 and 0.001 depending on the model and dataset, and decay it by 0.5 factor upon reaching a validation loss plateau. We limit the training time up to 60h, which is adequate for the models to converge, except SAN on \coco. SAN is particularly computationally intensive and may require a week of single NVidia A100 computation time to converge on \coco. Individual configuration files with exact hyperparameters for all 7 models and 5 datasets are provided with the source code.

\subsection{Computing environment and used resources}\label{app:cluster}
Our implementation uses GraphGPS~\cite{rampasek2022GPS} built on PyG and its GraphGym module \cite{FeyLenssen2019PyG,you2020design} that are all released under MIT License. All presented experiments were executed in a shared computing cluster environment (Digital Research Alliance of Canada and Mila Quebec AI Institute) with multiple CPU and GPU architectures: NVidia V100 (32GB), NVidia RTX8000 (48GB), and NVidia A100 (40GB). The resource budget for each experiment was 1 GPU, 4 CPUs, and up to 32GB system RAM. Except \coco, which required up to 72GB RAM. Average run times are shown in Table~\ref{tab:run-times}.

\begin{table}[ht]
\caption{Wall-clock run times. Average epoch time (average of 5 epochs, including validation performance evaluation) is shown for each model and dataset combination. Additionally, the precomputation time needed for LapPE and RWSE statistics is listed in the bottom of the table. The times were measured on a single NVidia A100 GPU system with 4 CPU cores of AMD Milan 7413.}\label{tab:run-times}
    \scalebox{0.85}{
    \centering
    \setlength\tabcolsep{4pt} 
    \rowcolors{2}{lightgray}{}
    \renewcommand{\arraystretch}{1.2}
    \begin{tabular}{lrrrrrr}\toprule
    \emph{avg. time / epoch} &\pascal &\coco &\pcqmcontact &\pepfunc & \pepstruct \\\midrule
    GCN &8.8s &111s &138s &3.0s &2.6s \\
    GCNII &8.2s &106s &137s &2.7s &2.4s \\
    GINE &7.2s &91s &138s &2.5s &2.6s \\
    GatedGCN &12s &151s &138s &3.3s &3.3s \\ \midrule
    Transformer+LapPE &13s &154s &145s &5.8s &5.9s \\
    SAN+LapPE &179s &2190s &793s &54.7s &53.6s \\
    SAN+RWSE &165s &2014s &740s &49.1s &49.7s \\ \midrule
    LapPE precomp. &8min 40s &1h 34min &5min 18s &1min 13s &1min 14s \\
    RWSE precomp. &7min 51s &1h 24min &6min 29s &53s &53s \\
    \bottomrule
    \end{tabular}
    }
\end{table}

\section{Additional Experiments with \texorpdfstring{$L=2$}{L=2} MP-GNNs} \label{app:2layer-mpnns}
\setcounter{figure}{0}
\setcounter{table}{0}

Here we investigate a shallow but wider variation of the baseline MP-GNN models. Instead of 5 or 8 layers (Table~\ref{tab:hparams}) we consider 2-layer MP-GNN architectures that allow for larger hidden node representations within the 500k parameter budget, Table~\ref{tab:hparams_2l}. These provide additional set of baselines, that investigate whether this limited receptive field of MP-GNNs is to a detriment in the proposed LRGB datasets, and also, whether the deeper architectures, evaluated in the main text, are not suffering from catastrophic over-smoothing and/or over-squashing. Note, that we do use residual connections in all our baseline models, this has been shown to significantly help to prevent deterioration of the models' performance with increasing depth.

Generally, we observe majorily decreased performance of 2-layer MP-GNNs as compared to their deeper versions, while their relative ordering by their performance remains largely the same. This finding confirms that the access to only a narrow receptive field is severely limiting. Additionally, we observe \emph{much increased positive impact} of augmenting 2-layer GatedGCN with positional or structural encodings. GatedGCN with LapPE or RWSE outperforms standard GatedGCN (and any other tested MP-GNN) by a large margin particularly in \pcqmcontact, \pepfunc, and \pepstruct. In the case of the deeper MP-GNN configurations (Table~\ref{tab:hparams}) this effect is not observed, suggesting that the positional or structural encodings provide additional information beyond the 2-hop neighborhood that a deeper GatedGCN appears to be able to substitute.

\begin{table}[ht]
    \caption{Hyperparameters of evaluated \emph{shallow} MP-GNN baseline models. The number of layers is set to $L$=2 and the size of the hidden node representation $d$ is set to fill 500k parameter budget. Where applicable, the type of positional/structural embedding is shown.}
    \label{tab:hparams_2l}
    \scalebox{0.85}{
    \centering
    \setlength\tabcolsep{4pt} 
    \rowcolors{2}{lightgray}{}
    \renewcommand{\arraystretch}{1.2}
    \begin{tabular}{l c c c c c}\toprule
    &\pascal &\coco &\pcqmcontact &\pepfunc & \pepstruct \\\midrule
    GCN &$d$=350, $L$=2 &$d$=345, $L$=2 &$d$=380, $L$=2 &$d$=460, $L$=2 &$d$=460, $L$=2 \\
    GCNII &$d$=350, $L$=2 &$d$=345, $L$=2 &$d$=380, $L$=2 &$d$=460, $L$=2 &$d$=460, $L$=2 \\
    GINE &$d$=285, $L$=2 &$d$=285, $L$=2 &$d$=300, $L$=2 &$d$=330, $L$=2 &$d$=330, $L$=2 \\
    GatedGCN(+PE/SE) &$d$=200, $L$=2 &$d$=200, $L$=2 &$d$=210, $L$=2 &$d$=215, $L$=2 &$d$=215, $L$=2 \\
    \hspace{3em} used PE/SE &LapPE-10 &LapPE-10 &RWSE-16 &RWSE-16 &RWSE-16 \\
    \bottomrule
    \end{tabular}
    }
\end{table}

\begin{table}[!ht]
    \caption{Baseline experiments for \pascal and \coco with \rbgraph graph on SLIC compactness 30 for node classification task for MP-GNNs with 2 layers and 500k parameters. Performance metric is macro F1 on the respective splits (Higher is better). All experiments are run 4 times with 4 different seeds. 
    }
    \label{tab:experiments_superpixels_additional}
    \begin{adjustwidth}{-2.5 cm}{-2.5 cm}\centering
    \scalebox{0.87}{
    \setlength\tabcolsep{4pt} 
    \begin{tabular}{l c A B r A B}
    \toprule
    \multirow{2}{*}{\textbf{Model ($L=2$)}} & \multirow{2}{*}{\hspace*{-1em}\textbf{\# Params.}} & \multicolumn{2}{c}{\pascal} & \multirow{2}{*}{\textbf{\# Params}} & \multicolumn{2}{c}{\coco} \\ \cmidrule(lr){3-4}\cmidrule(lr){6-7}
    & & \textbf{Train F1} & \textbf{Test F1 $\uparrow$} & & \textbf{Train F1} & \textbf{Test F1 $\uparrow$}\\
    \midrule 
    GCN & 504k & 0.1014$\pm$0.0031 & 0.1011$\pm$0.0024 & 511k & 0.0589$\pm$0.0019 & 0.0562$\pm$0.0015\\
    GCNII & 503k & 0.1137$\pm$0.0055 & 0.1067$\pm$0.0028 & 509k & 0.0705$\pm$0.0017 & 0.0656$\pm$0.0018\\
    GINE & 500k & 0.1467$\pm$0.0116 & 0.1238$\pm$0.0046 & 517k & 0.1110$\pm$0.0043 & 0.0929$\pm$0.0027\\
    GatedGCN & 491k & 0.2382$\pm$0.0313 & 0.2114$\pm$0.0157 & 503k & 0.1581$\pm$0.0033 & 0.1476$\pm$0.0027\\
    GatedGCN+LapPE & 492k & 0.2583$\pm$0.0458 & 0.2232$\pm$0.0255 & 504k & 0.1668$\pm$0.0037 & 0.1553$\pm$0.0026\\
    \bottomrule
    \end{tabular}
    }
    \end{adjustwidth}
\end{table}

\begin{table}[!ht]
    \caption{Baselines for \pepfunc (graph classification) and \pepstruct (graph regression) for MP-GNNs with 2 layers and 500k parameters. Performance metric is Average Precision (AP) for classification and MAE for regression. Each experiment was run with 4 different seeds.
    }
    \label{tab:experiments_peptides_additional}
    \begin{adjustwidth}{-2.5 cm}{-2.5 cm}\centering
    \scalebox{0.9}{
    \setlength\tabcolsep{4pt} 
    \begin{tabular}{l c A B A C}\toprule
    \multirow{2}{*}{\textbf{Model ($L=2$)}} &\multirow{2}{*}{\textbf{\# Params.}} &\multicolumn{2}{c}{\pepfunc} &\multicolumn{2}{c}{\pepstruct} \\ \cmidrule(lr){3-4} \cmidrule(lr){5-6}
    & &\textbf{Train AP} &\textbf{Test AP $\uparrow$} &\textbf{Train MAE} &\textbf{Test MAE $\downarrow$} \\\midrule
    GCN & 509k & 0.4956$\pm$0.0079 & 0.4566$\pm$0.0059 & 0.3836$\pm$0.0019 & 0.3950$\pm$0.0017\\
    GCNII & 507k & 0.5543$\pm$0.0077 & 0.4894$\pm$0.0039 & 0.3809$\pm$0.0020 & 0.3929$\pm$0.0020\\
    GINE & 501k & 0.5916$\pm$0.0189 & 0.5003$\pm$0.0042 & 0.3730$\pm$0.0029 & 0.3879$\pm$0.0011\\
    GatedGCN & 508k & 0.6085$\pm$0.0071 & 0.5073$\pm$0.0036 & 0.3757$\pm$0.0015 & 0.3905$\pm$0.0006\\
    GatedGCN+RWSE & 505k & 0.7946$\pm$0.0148 & 0.5812$\pm$0.0053 & 0.3173$\pm$0.0084 & 0.3599$\pm$0.0007\\ 
    \bottomrule
    \end{tabular}
    }
    \vspace{-8pt}
    \end{adjustwidth}
\end{table}

\begin{table}[!ht]
    \caption{Baseline performance on \pcqmcontact (link prediction) for MP-GNNs with 2 layers and 500k parameters. Each experiment was repeated with 4 different random seeds. 
    }
    \label{tab:experiments_pcqmcontact_additional}
    \begin{adjustwidth}{-2.5 cm}{-2.5 cm}\centering
    \scalebox{0.9}{
    \setlength\tabcolsep{3pt} 
    \begin{tabular}{l c B B B B}\toprule
    \textbf{Model ($L=2$)} &\textbf{\# Params.} &\textbf{Test Hits@1 $\uparrow$} &\textbf{Test Hits@3 $\uparrow$} &\textbf{Test Hits@10 $\uparrow$} &\textbf{Test MRR $\uparrow$} \\\midrule
    GCN & 500k & 0.0588$\pm$0.0007 & 0.1717$\pm$0.0011 & 0.5664$\pm$0.0008 & 0.1939$\pm$0.0003\\
    GCNII & 499k & 0.0651$\pm$0.0035 & 0.1714$\pm$0.0026 & 0.5399$\pm$0.0124 & 0.1944$\pm$0.0020\\
    GINE & 507k & 0.0596$\pm$0.0005 & 0.1718$\pm$0.0009 & 0.5685$\pm$0.0006 & 0.1949$\pm$0.0006\\
    GatedGCN & 528k & 0.0556$\pm$0.0008 & 0.1707$\pm$0.0014 & 0.5734$\pm$0.0027 & 0.1927$\pm$0.0010\\
    GatedGCN+RWSE & 525k & 0.1068$\pm$0.0010 & 0.3383$\pm$0.0012 & 0.8036$\pm$0.0008 & 0.2937$\pm$0.0007\\ 
    \bottomrule
    \end{tabular}
    }
    \end{adjustwidth}
\end{table}

\newpage
\section{Inspection of Transformer attention} \label{app:attn-analysis}
\setcounter{figure}{0}
\setcounter{table}{0}

In this section we investigate how a Transformer with LapPE \cite{dwivedi2020generalization} processes the 5 proposed LRGB datasets and an existing MNIST dataset~\cite{dwivedi2020benchmarking}. In particular, we investigate how strongly a Transformer attends to nodes that are at various $k$ distances away from a node $v$ during updating of its representation $h_v^\ell$ at layer $\ell \in \{0, \dots, L-1\}$. The goal is to probe whether a model capable of global attention, such as the Transformer with LapPE, in fact attends to nodes farther than the local neighborhood of $v$, while performing better or comparable to local MP-GNN models.

\begin{figure}[ht]
    \centering
    \includegraphics[trim=0 0 0 0, width=1\textwidth]{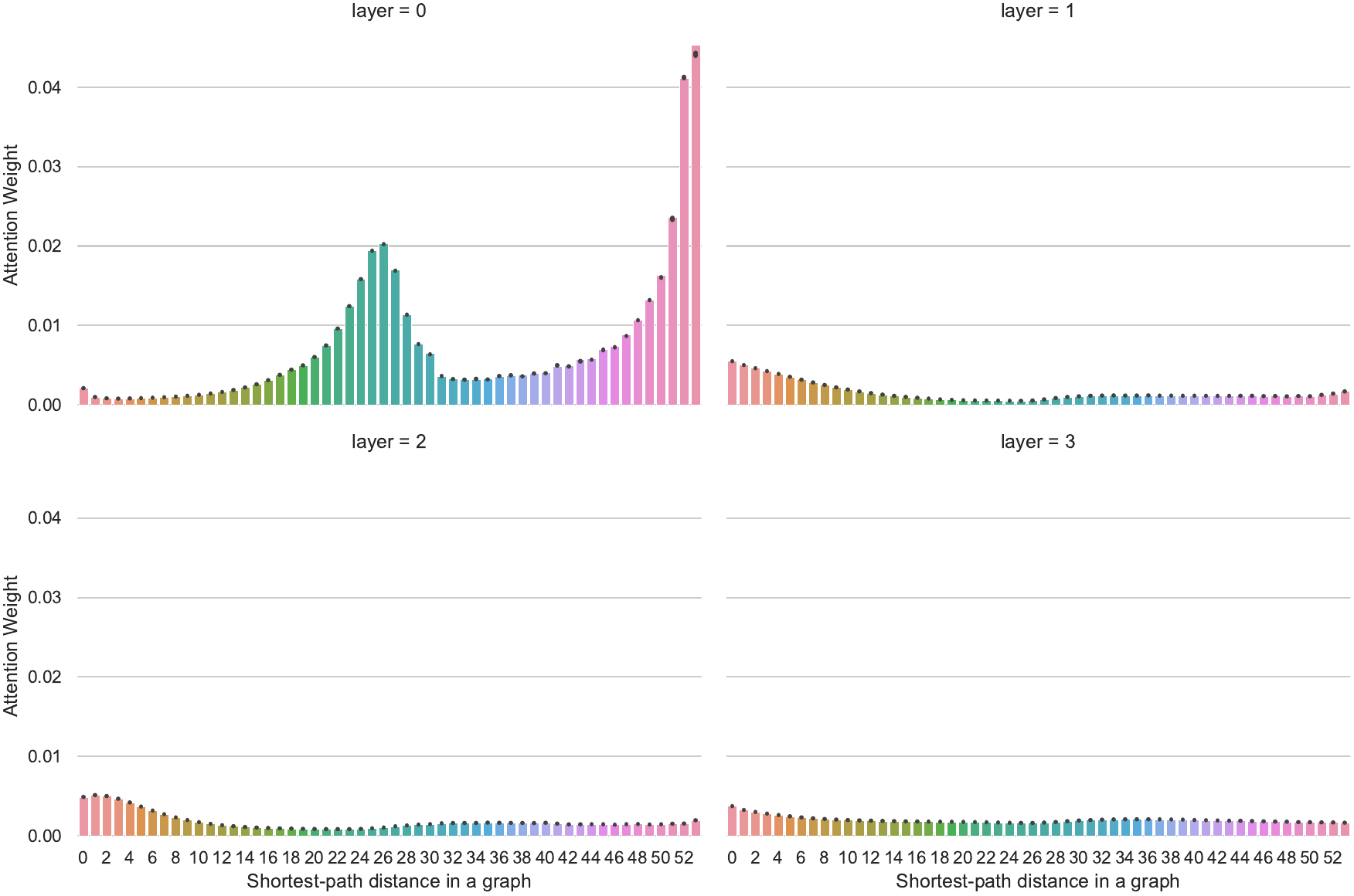}
\caption{Average attention weight distribution of Transformer+LapPE on \pascal dataset.}
\label{fig:attn-pascal}
\end{figure}

\begin{figure}[ht]
    \centering
    \includegraphics[trim=0 0 0 0, width=1\textwidth]{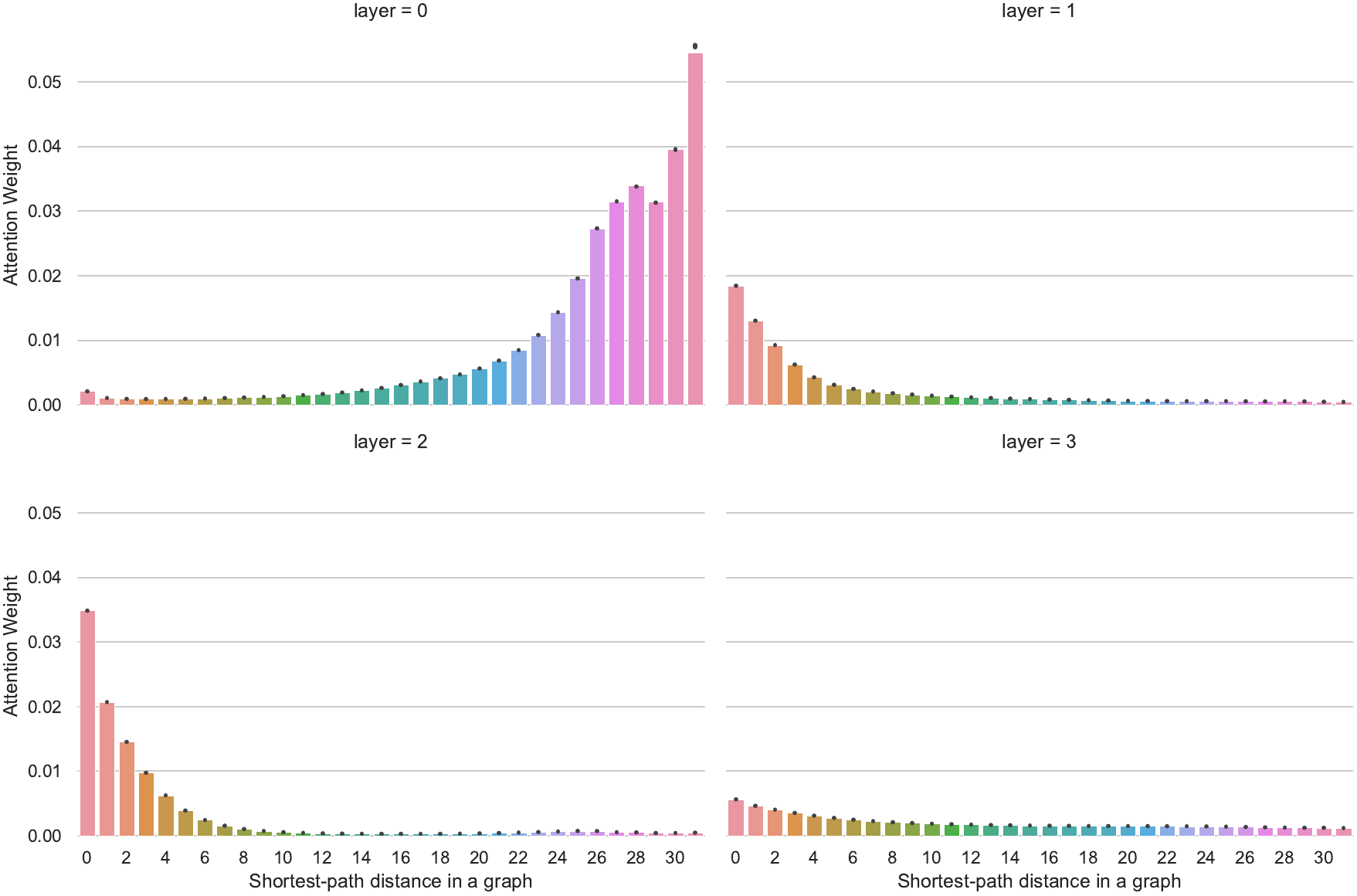}
\caption{Average attention weight distribution of Transformer+LapPE on \coco dataset.}
\label{fig:attn-coco}
\end{figure}

\begin{figure}[ht]
    \centering
    \includegraphics[trim=0 0 0 0, width=1\textwidth]{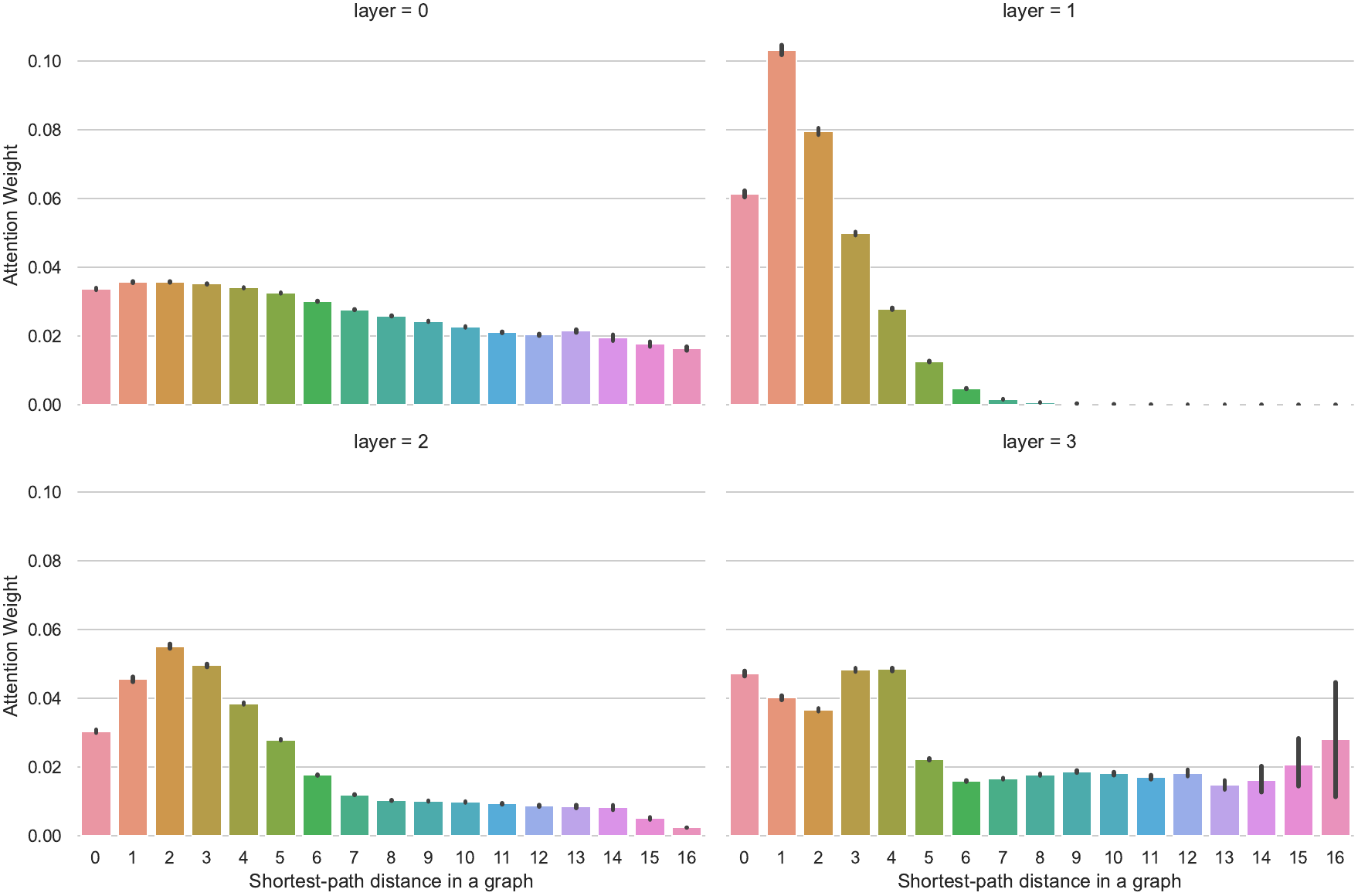}
\caption{Average attention weight distribution of Transformer+LapPE on \pcqmcontact dataset.}
\label{fig:attn-pcqmcontact}
\end{figure}

\begin{figure}[ht]
    \centering
    \includegraphics[trim=0 0 0 0, width=1\textwidth]{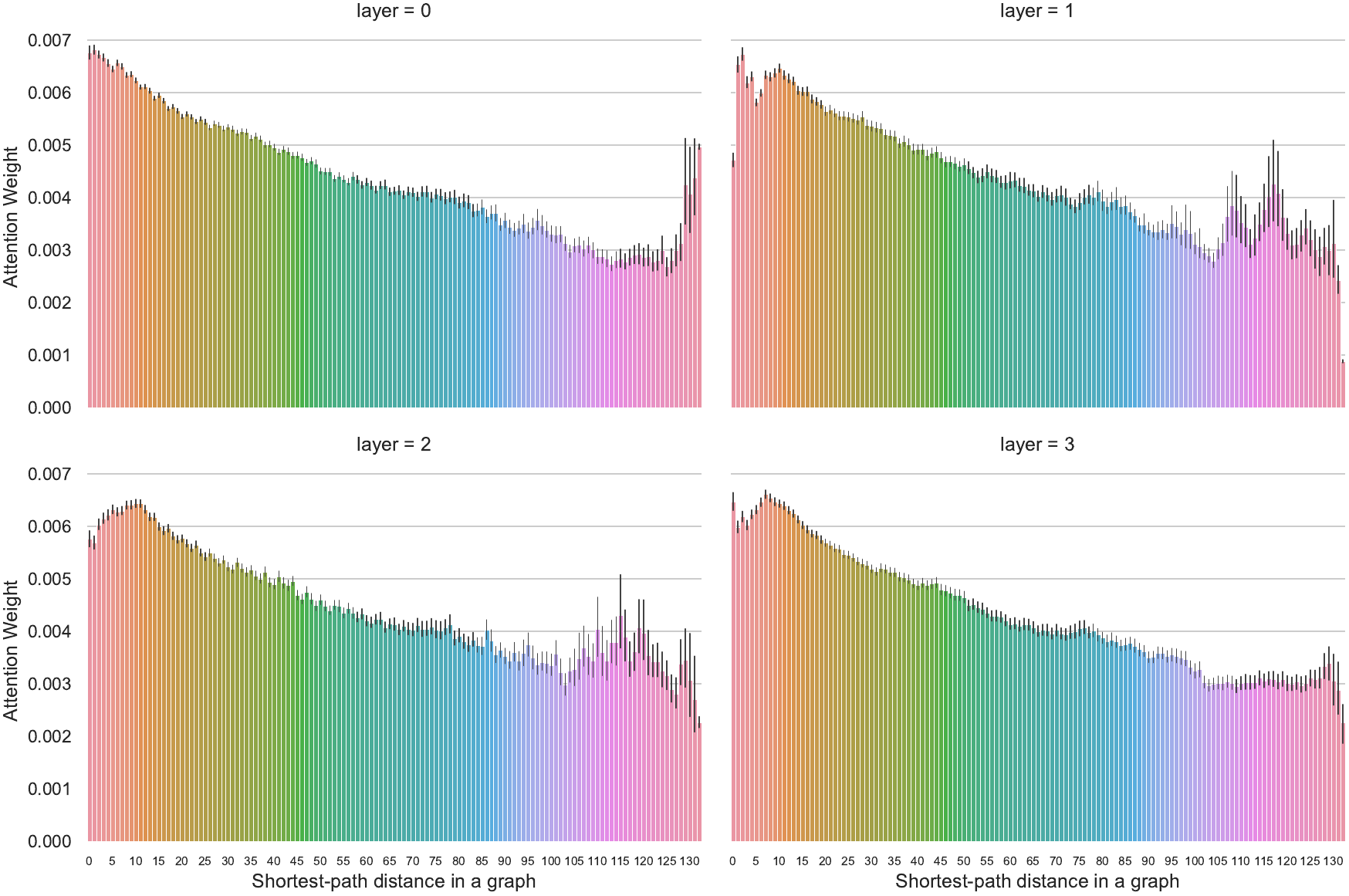}
\caption{Average attention weight distribution of Transformer+LapPE on \pepfunc.}
\label{fig:attn-peptides-func}
\end{figure}

\begin{figure}[ht]
    \centering
    \includegraphics[trim=0 0 0 0, width=1\textwidth]{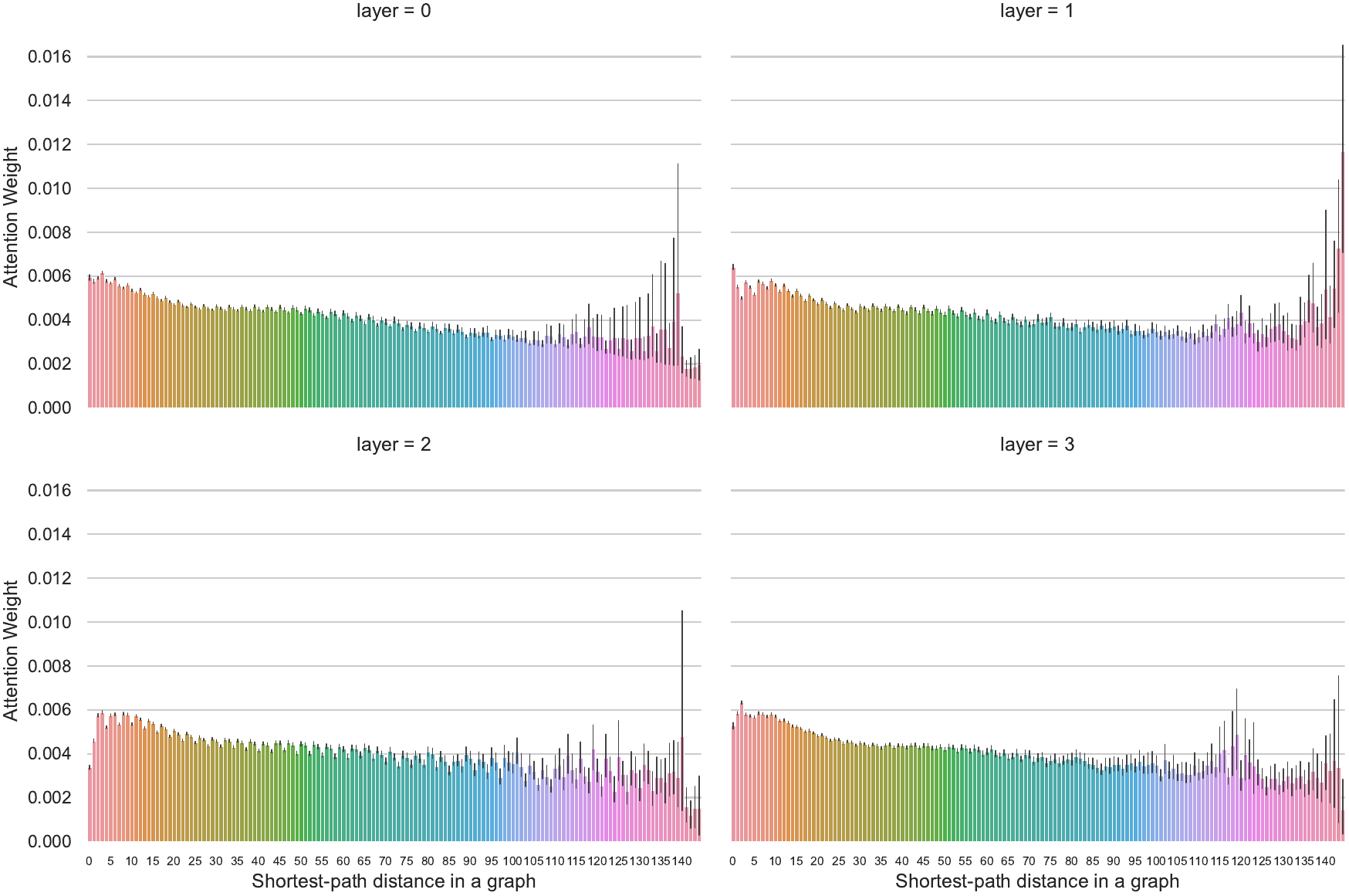}
\caption{Average attention weight distribution of Transformer+LapPE on \pepstruct.}
\label{fig:attn-peptides-struct}
\end{figure}

For each dataset, we used a fully trained graph Transformer model with LapPE (using the same hyperparameters and training pipeline as described in Appendix~\ref{app:exp_details}) and inspected its attention weights on 128 randomly selected test graphs. For each layer $\ell$, we plot the average attention weight aggregated by how far in the graph the node is from the perspective of a ``focal'' node $v$ that is being processed. That is,  we show what attention weight on average a node $u$ that is $k$-hops away from $v$ (shortest-path distance $k$) gets. Note, that attention to a node at distance $k=0$ denotes the attention of $v$ to self. The resulting bar plots of attention weights in each of the 5 LRGB datasets are shown in Figures~\ref{fig:attn-pascal}-\ref{fig:attn-peptides-struct}.

Overall the attention distributions vary across datasets and layers, but generally confirm that Transformer exhibits attention patterns beyond local neighborhoods. In \pascal and \coco the first layer ($\ell = 0$) shows higher attention to mid- and long-distance nodes over the close-by nodes; this changes in the second and third layers, where attention to close-by nodes is dominant; and finally the last layer exhibits the most even attention distribution with some bias towards close-by nodes. In \pcqmcontact dataset, the attention distributions are similar, except the first layer ($\ell = 0$) that is much more uniform yet lightly favoring nodes in the first half of the distance range. Finally, in \pepfunc and \pepstruct the attention distributions are considerably more consistent across the layers and exhibit mostly linear attention weight decay with the growing shortest-path distance between the nodes.

\begin{figure}[ht]
    \centering
    \includegraphics[trim=0 0 0 0, width=1\textwidth]{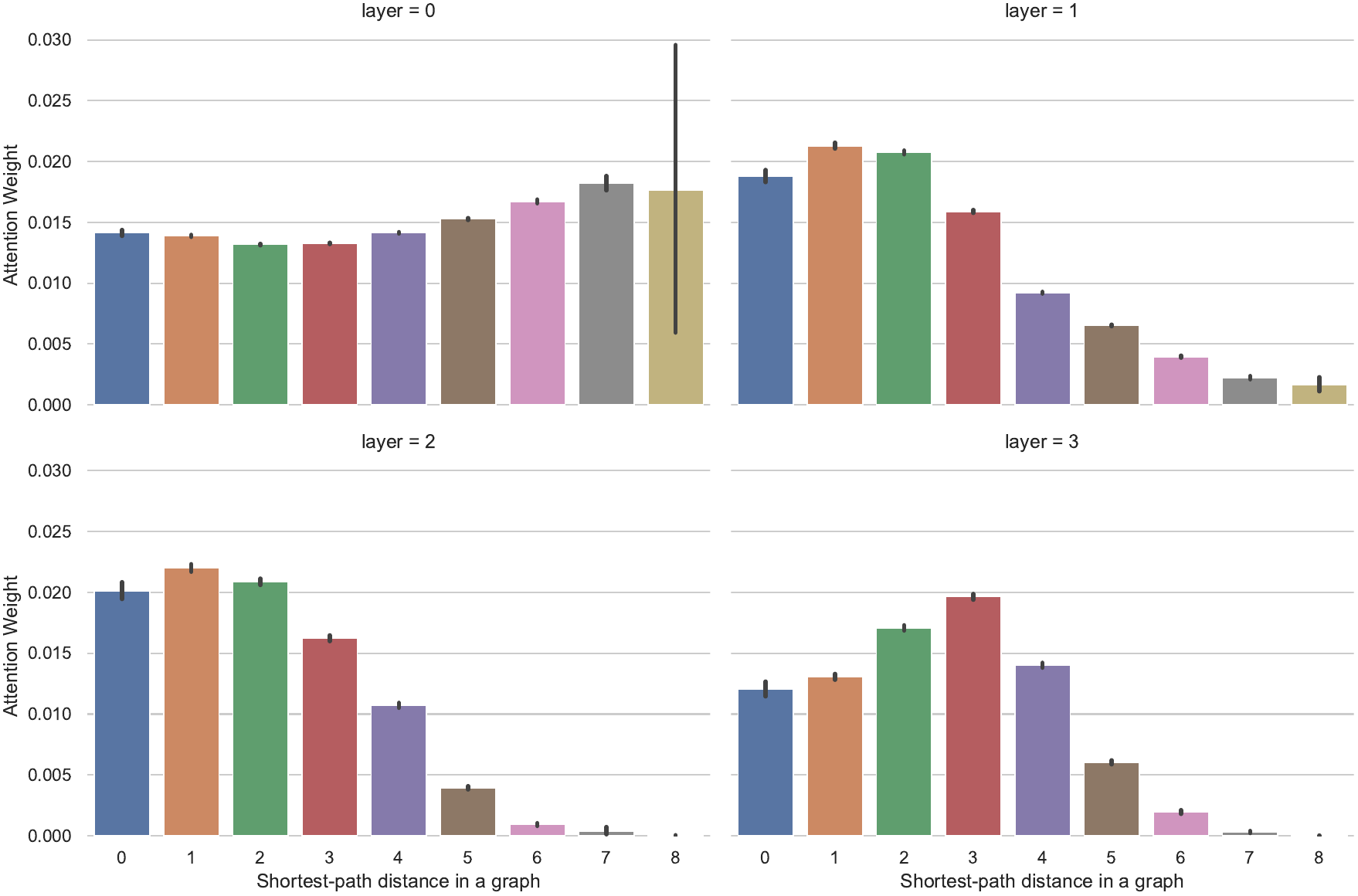}
\caption{Average attention weight distribution of Transformer+LapPE on MNIST dataset~\cite{dwivedi2020benchmarking}. Compared to the proposed 5 LRGB datasets, graphs in MNIST dataset have much smaller graph diameter and except the first layer (layer = 0), the attention is majorly focused on close neighbors that are up to 4-hops away.}
\label{fig:attn-mnist}
\end{figure}

In addition to the above LRGB datasets, we conducted the same inspection of a Transformer model with LapPE on an existing dataset MNIST~\cite{dwivedi2020benchmarking}, that we argue is insufficient for benchmarking a model's ability to capture LRIs. We stick to 100k parameter budget, as per standard for this dataset, using 4 layers with hidden node representation size of 52. A Transformer+LapPE model scores $97.89\%$ test accuracy on a random split, and its attention distribution is shown in Figure~\ref{fig:attn-mnist}. Graphs in MNIST dataset have much smaller graph diameter and except for the first layer ($\ell = 0$), the attention is majorly focused on close neighbors that are up to 4-hops away. While on its own a not sufficient proof, the difference in attention distributions between LRGB datasets and MNIST support the viability of proposed LRGB datasets for testing a model's capability to capture interactions beyond limited local neighborhoods.

\end{document}